\documentclass[letterpaper, 10 pt, journal, twoside]{URL-ieeeconf}  

\usepackage{graphics}           
\usepackage{soul}
\usepackage{times}            
\usepackage{amsmath}            
\usepackage{amssymb}            
\usepackage{graphicx}
\usepackage{algorithm}
\usepackage[noend]{algpseudocode}
\usepackage{booktabs}
\usepackage{color}
\usepackage{multirow}
\usepackage{subcaption}
\usepackage{rotating}
\usepackage{cite}
\usepackage{tikz}
\usepackage{pifont}
\usepackage{xspace}
\usepackage{xcolor}

\usepackage{array}
\usepackage{verbatim} 
\makeatletter
\let\NAT@parse\undefined
\makeatother
\usepackage{hyperref}
\usepackage{cleveref}

\definecolor{instructioncolor}{rgb}{.0,.0,.0}
\definecolor{instructioncolor2}{rgb}{.0,.0, .0}
\definecolor{resubmitcolor}{rgb}{.0,.0, 0.0}
\definecolor{profcolor}{RGB}{235, 100, 36}
\definecolor{darkred}{rgb}{.804,.196,.196}
\definecolor{darkorange}{rgb}{1.0,.55,.0}
\definecolor{darkgreen}{rgb}{.196,.70,.196}

\newcommand{\revision}[1]{\textcolor{instructioncolor2}{#1}}

\newcommand{\changed}[1]{\textcolor{instructioncolor}{#1}}
\newcommand{\resubmit}[1]{\textcolor{resubmitcolor}{#1}}

\newcommand*{\acro}{Chamelion\xspace}
\newcommand{\hatop}{\hat{(\cdot)}} 

\def\secref#1{Section~\ref{#1}}
\def\figref#1{Fig.~\ref{#1}}

\def\tabref#1{Table~\ref{#1}}
\def\eqref#1{(\ref{#1})}

\def\psdscan{\tilde{\mathcal{S}}_{0:{T}}}
\def\psdonescan{\tilde{\mathcal{S}}_{i}}
\def\psdmap{\tilde{\mathcal{M}}}

\def\vsfigu{\vspace{-0.1cm}}
\def\vsfiguu{\vspace{0.1cm}}
\def\vsfig{\vspace{-0.3cm}}

\def\vsequ{\vspace{-0.15cm}}

\newcommand{\rom}[1]{\uppercase\expandafter{\romannumeral #1\relax}}

\makeatletter
\DeclareRobustCommand\onedot{\futurelet\@let@token\@onedot}
\def\@onedot{\ifx\@let@token.\else.\null\fi\xspace}

\def\etal{{\textit{et al}}\onedot}
\makeatother

\def\etalcite#1{\etal~\cite{#1}}

\newcolumntype{L}[1]{>{\raggedright\let\newline\\\arraybackslash\hspace{0pt}}m{#1}}
\newcolumntype{C}[1]{>{\centering\let\newline\\\arraybackslash\hspace{0pt}}m{#1}}
\newcolumntype{R}[1]{>{\raggedleft\let\newline\\\arraybackslash\hspace{0pt}}m{#1}}

\sethlcolor{lightgray}

\title{\LARGE \bf Chamelion: Reliable Change Detection for \\ Long-Term LiDAR Mapping in Transient Environments}

\author{Seoyeon Jang$^1$, Alex Junho Lee$^2$, I Made Aswin Nahrendra$^3$, and Hyun Myung$^{1*}$, \textit{Senior Member, IEEE} 
\thanks{$^*$Corresponding author: Hyun Myung}
\thanks{$^{1}$Seoyeon Jang and Hyun Myung are with the School of Electrical Engineering, KAIST (Korea Advanced Institute of Science and Technology), Daejeon, 34141, Republic of Korea. {\tt\scriptsize \{9uantum01, hmyung\}@kaist.ac.kr}}
\thanks{$^{2}$Alex Junho Lee is with the Department of Electrical Mechanical Systems Engineering, Sookmyung Women's University, Seoul, 04310, Republic of Korea. {\tt\scriptsize alexlee@sookmyung.ac.kr}}
\thanks{$^{3}$I Made Aswin Nahrendra is with KRAFTON,~Seoul,~06142,~Republic of Korea.~{\tt\scriptsize anahrendra@krafton.com}
\hfill \break
}
}

\begin{document}


\maketitle


\begin{abstract}
Online change detection is crucial for mobile robots to efficiently navigate through dynamic environments. 
Detecting changes in transient settings, such as active construction sites or frequently reconfigured indoor spaces, 
is particularly challenging due to frequent occlusions and spatiotemporal variations. 
Existing approaches often struggle to detect changes and fail to update the map across different observations. 
To address these limitations, we propose a dual-head network designed for online change detection and long-term map maintenance. 
A key difficulty in this task is the collection and alignment of real-world data, 
as manually registering structural differences over time is both labor-intensive and often impractical. 
To overcome this, we develop a data augmentation strategy that synthesizes structural changes by importing elements from different scenes, 
enabling effective model training without the need for extensive ground-truth annotations. 
Experiments conducted at real-world construction sites and in indoor office environments demonstrate that 
our approach generalizes well across diverse scenarios, achieving efficient and accurate map updates.
\resubmit{Our source code and additional material are available at:~\href{https://chamelion-pages.github.io/} {https://chamelion-pages.github.io/}.} 
\end{abstract}

\begin{keywords}
  Mapping; Object Detection, Segmentation and Categorization
\end{keywords}
\vspace{-0.3cm}

\section{Introduction}\label{sec:introduction}
\PARstart{O}{nline} change detection is the task of identifying structural discrepancies during robot operation between a previously constructed map and the current sensor observation. 
In long-term environmental monitoring or inspection, this capability is crucial for maintaining an accurate understanding of the environment, 
especially in dynamic settings such as ongoing construction sites or disaster-affected areas. 
Unlike open environments where main structures generally remain visible, major robot service spaces such as crowded indoor environments involve moving objects, frequent occlusions, and significant structural changes. 
Online map alignment is particularly important in these contexts, 
especially when multiple robots are sharing information or when a robot revisits a location after an extended period. 
In such cases, aligning current observations with previously collected data is necessary to ensure consistent localization and mapping across different times and agents. 
To properly operate in these applications, spatiotemporal changes must be recognized at scan time, 
enabling robots to dynamically adjust their behavior in response to the evolving environment.

\begin{figure}[t!]
	\captionsetup{font=footnotesize}
	\centering
	\includegraphics[width=0.47\textwidth]{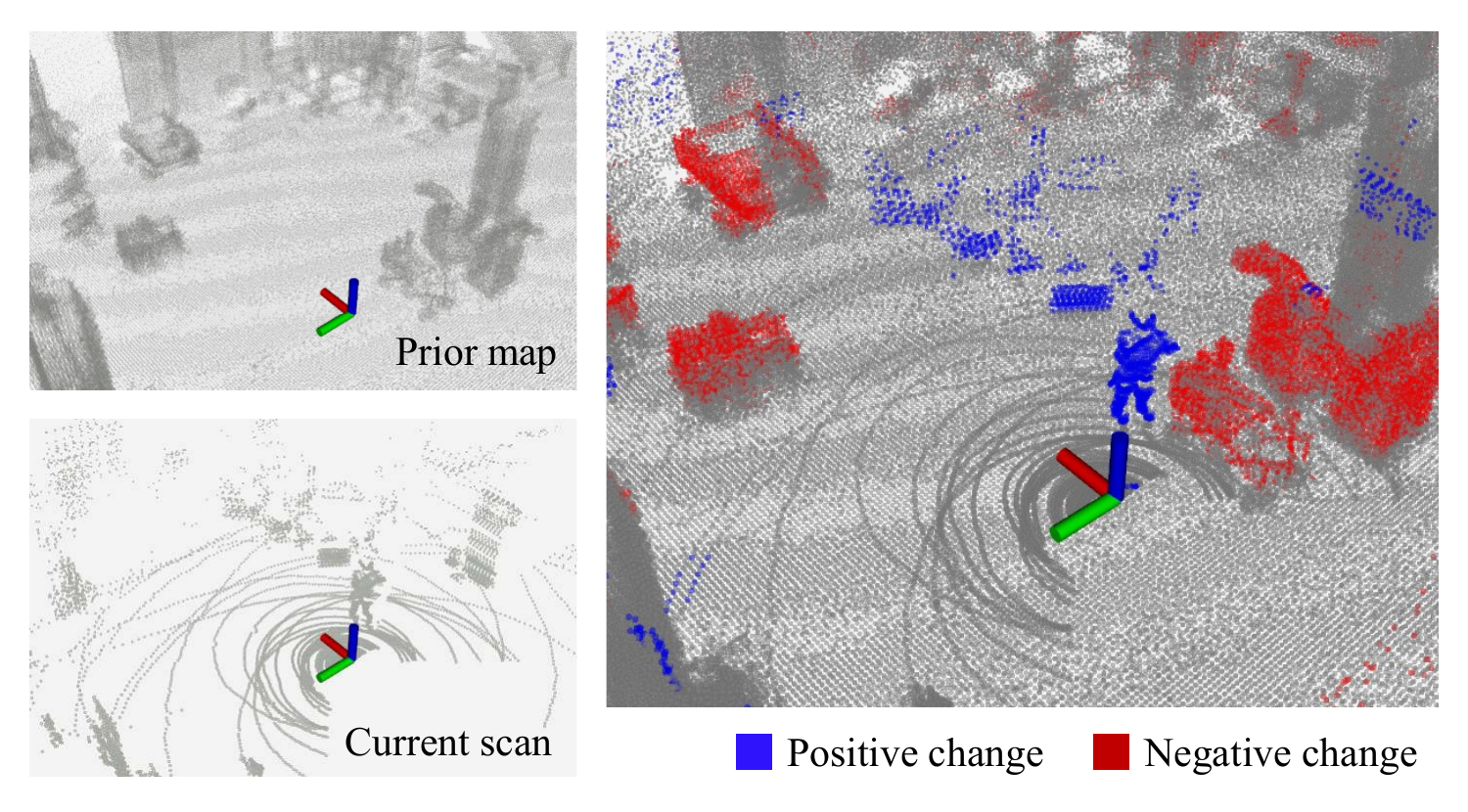}
	\caption{Our method detects low dynamic~(LD) changes in real-time between the prior map and the current scan. 
	Here, red points represent negative changes~\changed{(NC)} that have disappeared from the map, while blue points indicate positive changes~\changed{(PC)} that have newly appeared in the current scan.}
	\label{fig:main_figure}
	\vsfig
\end{figure}

Changes can be categorized into two types based on their behavior within the period of a single observation: high-dynamic (HD) changes and low-dynamic (LD) changes.
HD changes refer to inconsistencies that occur during the observation window, such as walking pedestrians or moving vehicles, 
whereas LD changes correspond to structural modifications that remain static within a single scan but differ across sessions, as illustrated in \figref{fig:main_figure}.
LD changes can be further classified into positive changes~(PC), where new structures appear, and negative changes (NC), where previously existing elements disappear.

While extensive research has been conducted on HD change detection~\cite{lim2021ral, lim2023erasor2, mersch2022ral,chen2021ral}, 
LD change detection has received limited attention primarily because it hardly affects local odometry, and moreover, it is difficult to analyze.
As relocalization becomes less critical with accurate odometry, prior work has largely focused on handling HD changes.
However, even after removing HD changes, unlabeled LD changes can drive the map unreliable when observations are accumulated over multiple sessions, 
degrading map quality for localization.
To address this problem, we propose a method to detect LD changes and update the prior map to maintain long-term consistency.

LD change detection has been studied using both 2D images~\cite{sakurada2020icra, chen2021iv, park2022cvpr} and 3D range sensors~\cite{walcot2012iros, fehr2017icra, schmid2022icra, underwood2013icra, kim2022icra}.
Among these, 3D sensors are particularly suitable for capturing structural information and supporting reliable map management in dynamic environments.
Classical 3D range sensor-based change detection methods primarily rely on geometric differences between two observations.
However, geometry-based methods are generally more vulnerable to occlusion. Moreover, despite potential parallelization,
classical methods still suffer from scalability issues as the environment size increases, mainly due to the growing number of points that must be compared.
Recently, deep learning-based approaches have been actively explored for faster inference than classical methods~\cite{joseph2024icra,ku2021cg,hroob2024ias,hroob2024ral,krawciw2023arxiv}.
However, they still face fundamental challenges, particularly in long-term settings, 
including (1)~the difficulty of collecting large-scale training datasets that reflect long-term differences, 
and (2)~the challenge of distinguishing between occlusions and true changes.

In this paper, we address these challenges by designing a learning framework that explicitly separates occlusion and disappearance, 
and by constructing a data generation pipeline tailored to long-term environmental variations. 
Based on this idea, we propose \textbf{\acro}, {a novel framework for \textbf{Cha}nge detection and long-term \textbf{M}ap management in transient \textbf{E}nvironments, 
using \textbf{Li}DAR and designed for \textbf{On}line operation.}
Our main contributions can be summarized as follows:

First, to address the difficulty of obtaining large-scale long-term change datasets, we propose a composition-based augmentation strategy that synthetically generates pseudo-changes from single-session scans.
Second, we design a novel network architecture that employs a 4D CNN backbone~\cite{choy20194d} and a dual-head structure, 
consisting of a class head for change classification and a confidence head for occlusion awareness, to enhance generalization and robustness.
Finally, through extensive experiments on real-world and synthetic datasets, 
we show that our method outperforms existing approaches in change detection performance and enables effective online map updates.

\section{Related Works}\label{sec:related_works}

\subsection{Change Detection Dataset Generation}\label{subsec:change_dataset}

Generating 3D change detection datasets is challenging due to the need for multiple observations of the same environment over time and accurate change annotations.
Manual data collection and labeling, typically performed in small indoor environments, become impractical for large-scale environments such as construction sites~\cite{langer2020iros, fehr2017icra, wald2019iccv}.
To address these limitations, simulation-based datasets have been introduced. 
For example, Park~\etalcite{park2021iros} proposed the ChangeSim dataset for industrial indoor environments, and Joseph~\etalcite{joseph2024icra} presented 
the LiSTA dataset featuring LiDAR scans in office spaces. However, even simulation-based datasets still require multiple acquisitions of the same environment to simulate temporal differences.
Data synthesis techniques have been explored in the 2D domain to overcome this challenge by generating change detection datasets from a single image.
Park~\etalcite{park2022cvpr} improved 2D change detection by creating synthetic samples through random warping and cut-and-paste operations.
Inspired by these techniques, we propose a {composition}-based augmentation strategy that enables the generation of pseudo-change datasets using only single-session LiDAR scans, eliminating the need for multiple observations over time.
\vspace{-0.2cm}
\subsection{LiDAR-based Change Detection}\label{subsec:lidar_change_detection}

Traditional 3D change detection methods primarily rely on geometric differences across different sessions, such as occupancy changes~\cite{walcot2012iros}, TSDF differences~\cite{fehr2017icra, schmid2022icra} or visibility changes~\cite{underwood2013icra, kim2022icra}. However, {t}hey are sensitive to environmental variations, and their computational cost increases significantly as the number of point clouds grows. Recently, deep learning-based approaches aim to address some of this scalability issues. Supervised learning approaches rely on labeled datasets, but generating large-scale annotated data is time-consuming and labor-intensive\cite{ku2021cg}. Self-supervised methods have been proposed to reduce label dependency\cite{hroob2024ias}, yet they still require multiple observations of the same environment. 
{The~other} approach~\cite{krawciw2023arxiv} attempts to improve efficiency by converting 3D LiDAR data into 2D range images as network inputs, enabling real-time performance. However, it heavily relies on omnidirectional LiDAR, and it is less generalizable to other types of range sensors or diverse environments. Furthermore, occlusion handling remains an open challenge, requiring additional pre-processing or post-processing steps~\cite{joseph2024icra}, which increase pipeline complexity.
In this paper, we propose a novel LiDAR-based online change detection framework that explicitly separates true changes from occlusions, generalizes across different types of 3D range sensors, and eliminates the need for extensive pre- or post-processing.

\begin{figure}[t!]
    \centering
    \includegraphics[width=0.45\textwidth]{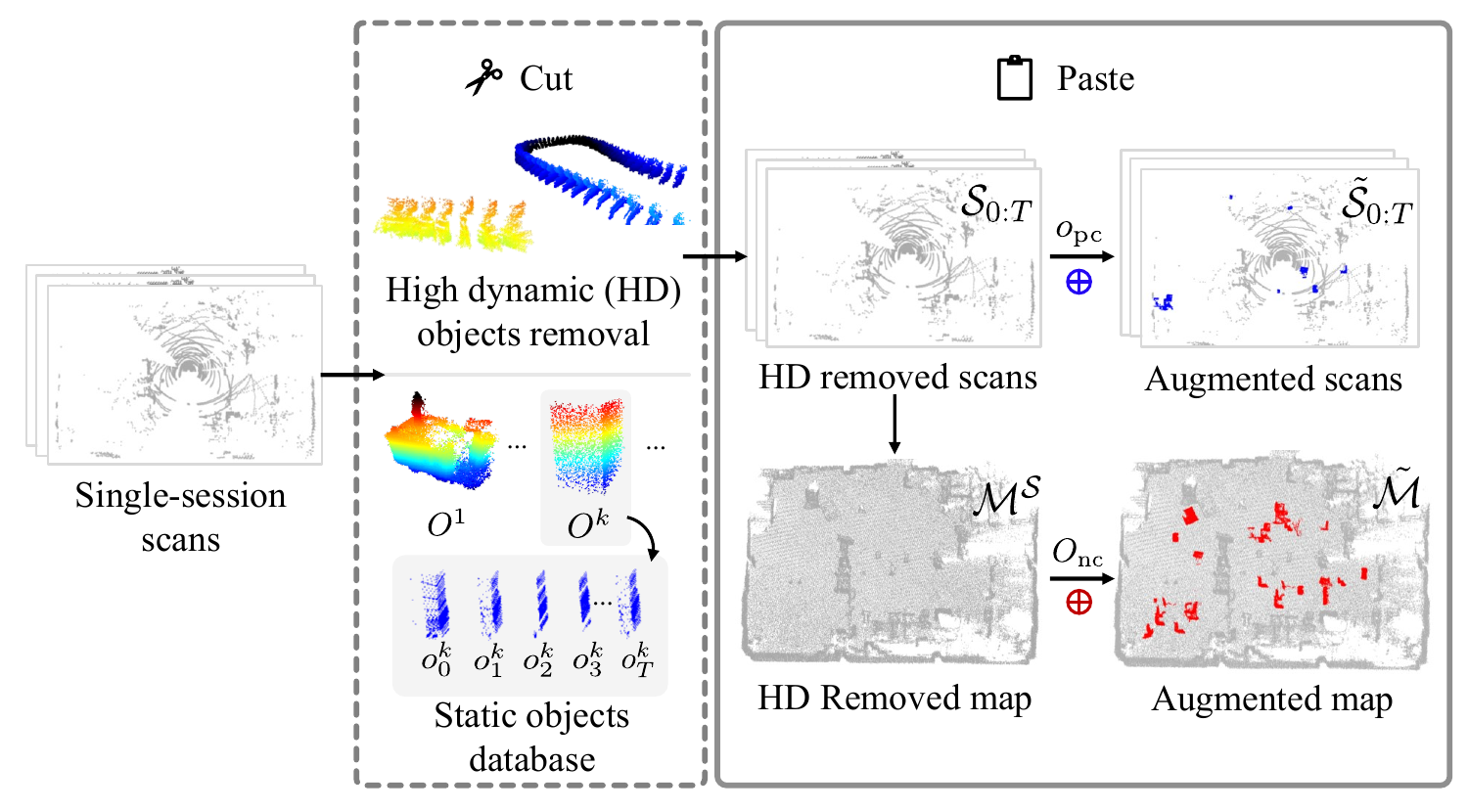}
    \captionsetup{font=footnotesize}
    \caption{Our proposed composition-based augmentation for the pseudo-label generation. 
    Static objects are extracted using multi-object tracking-based segmentation, where $O^k = \{o^k_j \mid j = {0, \dots, T}\}$ denotes snapshots of the $k$-th object.
    These objects are pasted onto arbitrary locations in single-session scans~($\mathcal{S}_{0:t}$) and the map~($\mathcal{M^S}$)
    to form a pseudo multi-session dataset, $\psdscan$ and $\psdmap$.
    Here, $o_\mathrm{pc}$ is inserted into a scan, and $O_\mathrm{nc}$ into the map.}
    \label{fig:mono_temporal_data_generation}
    \vspace{-0.5cm}
\end{figure}
\begin{figure*}[t!]
    \centering
    \begin{subfigure}[b]{0.98\textwidth}
		\centering
		\includegraphics[width=1.0\textwidth]{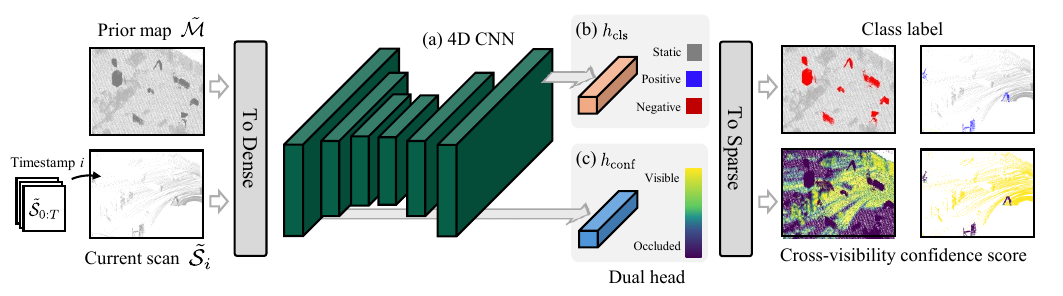}
	\end{subfigure}
    \vsfigu
    \captionsetup{font=footnotesize}
    \caption{Our dual-head architecture for change detection. {(a)}~First, we extract features from the input using a 4D convolutional neural network. 
    These features are then fed into two separate heads: one for {(b)}~change classification and the other for {(c)}~cross-visibility confidence estimation. 
    When updating the map, we only trust the class output in areas with high cross-visibility confidence, where both the map and scan are visible.
    }
    \label{fig:dual_head_architecture}
    \vsfig
\end{figure*}

\section{Our Approach for Online Change Detection}\label{sec:methodology}
To address the challenges in long-term LiDAR-based change detection, we propose a novel framework that
(1) generates training data from a single session without~requiring multiple temporal observations,
(2) explicitly~models occlusion uncertainty through cross-visibility confidence estimation, and
(3) updates the prior map probabilistically based on confidence-aware change detection.

This section details each component of our framework:
First, we introduce a composition-based augmentation strategy to generate pseudo-change data from a single session~(\secref{subsec:mono_temporal_datagen}).
Next, we describe our 4D sparse convolutional backbone and dual-head structure, 
which jointly predict changes and estimate cross-visibility confidence~(Sections~\ref{subsec:backbone}~and~\ref{subsec:cls_head}).
Finally, we present a confidence-aware, probabilistic update scheme for online map maintenance~(\secref{subsec:prob_map_update}).

\subsection{\changed{Composition-based Data Augmentation for Single-Session Pseudo-Label Generation}}\label{subsec:mono_temporal_datagen}
Training change detection requires multi-session data, which is costly and impractical to collect. 
To address this, we propose a single-session augmentation method that synthetically generates change pseudo-labels using only a single traversal of the environment, as illustrated in \figref{fig:mono_temporal_data_generation}. 
We start by constructing a static map from single-session LiDAR scans.

Let $\mathcal{S}_i^\text{local}$ denote the $i$th LiDAR scan at timestamp $t_i$ in the sensor coordinate frame, 
and $\mathcal{S}_i$ denote its transformation into the global coordinate frame with its estimated pose $G_i$,~such that~{$\mathcal{S}_i = G_i * \mathcal{S}_i^\text{local}$}, 
where $*$ denotes the transformation operation applied to all points of $\mathcal{S}_i^\text{local}$.
The global map $\mathcal{M}^\mathcal{S}$ is then constructed by taking the union of all transformed scans up to time $t_{T}$:
\begin{equation}
  \mathcal{M}^\mathcal{S} = \bigcup_{i=0}^{T} \mathcal{S}_i.
\end{equation}

To ensure that the map reflects only the static elements of the environment, \revision{we apply multi-object tracking-based moving object segmentation (MOS)~\cite{jang2023toss} to filter out high-dynamic~(HD) objects associated with the same ID, 
while retaining static objects with consistent IDs across scans.}
After filtering out HD objects, we save the point cloud of the $j$-th snapshot of the $k$-th static object, denoted as {$o_j^k$}~$(j = 0, \dots, T)$, 
and construct a set of static object snapshots $O^k$. Finally, the collection of all static object snapshots is stored as a database $\mathcal{DB}$, defined as:
\begin{align}
  O^k &= \{o^k_j \mid j = {0, \dots, T}\}, \\
  \mathcal{DB} &= \{O^k \mid k = {1, \dots, N_{T}}\},
  \end{align}
where $N_{{T}}$ is the number of static objects tracked up to time $t_{T}$.

Using this database, we generate training samples with change pseudo-labels by randomly sampling static objects and inserting them into scans and maps.
\revision{Prior to object insertion, ground segmentation~\cite{oh2022travel} is performed on each LiDAR scan, and insertion locations are preferentially selected from ground points to ensure geometrically plausible placement.}
Since the map is a dense accumulation of multiple scans, while each individual scan remains sparse, the insertion strategy differs between the two.
Specifically, we insert only single-frame object instances ($o^k_j$) into scans, and all accumulated snapshots of a tracked object ($O^k$) into maps.
Pseudo-labels are then assigned based on object placement:
objects added to the scan but absent from the map are labeled as positive changes~(\changed{\(o^k_j \rightarrow o_\mathrm{pc}\), \(O^k \rightarrow O_\mathrm{pc}\)}),
whereas objects added to the map but absent from the scan are labeled as negative changes~(\changed{\(o^k_j \rightarrow o_\mathrm{nc}\), \(O^k \rightarrow O_\mathrm{nc}\)}).
By expressing the scans accumulated up to timestamp $t_{T}$ as $\mathcal{S}_{0:{T}} = \{\mathcal{S}_i \mid i = 0, \dots, T\}$, 
the augmented map~($\psdmap$) and scans~($\psdscan$) with randomly placed objects are defined as:
\begin{align}
  \psdmap &= \mathcal{M}^\mathcal{S} \oplus O_\mathrm{\changed{nc}}, \\
  \psdscan &= \{\mathcal{S}_i \oplus o_\mathrm{\changed{pc}} \mid o_\mathrm{\changed{pc}} \in O_\mathrm{\changed{pc}}, \changed{i = 0, \dots, T}\}, 
\end{align}
where the operator $\oplus$ denotes object insertion into a scan or a map. 
To avoid any overlap between inserted objects, we assume $O_\mathrm{\changed{pc}} \cap O_\mathrm{\changed{nc}} = \varnothing$.
\changed{As a result, $\psdmap$ and $\psdscan$ serve as synthetic multi-session data, enabling effective training of change detection models.}

\subsection{4D Sparse Convolutional Backbone for \changed{Changed} Feature Extraction}\label{subsec:backbone}
To effectively capture both spatial and temporal features in sparse LiDAR data, 
we employ a 4D sparse convolutional neural network based on the MinkowskiEngine~\cite{choy20194d}.

The map $\psdmap \in \mathbb{R}^{m \times 3}$ and the {scan~$\psdonescan \in \mathbb{R}^{n \times 3}$, sampled from~$\psdscan$}, 
are labeled with a visibility flag $\nu$:~0 for all map points and 1 for all scan points.
Here, $m$ and $n$ are the total number of points from the map and scan, respectively. 
As a result, the two point sets are extended to tensors of size \(m \times 4\) and \(n \times 4\), 
where each point is represented as a 4D vector of $(x, y, z)$ coordinates and \changed{a visibility flag~$\nu$}.

The two sets are then concatenated into a 4D dense tensor:~$\mathcal{T} \in \mathbb{R}^{(m+n) \times 4}$.
This dense tensor is then discretized through a quantization process, 
resulting in a \changed{4D sparse tensor~$\mathcal{T}^{\prime} \in \mathbb{R}^{(m'+n') \times 4}$, 
where $m'$ and $n'$ represent the number of non-empty voxels corresponding to the map and scan points, respectively.}
The sparse tensor serves as the input to the backbone network.
The network outputs a set of feature representations, denoted as~$F \in \mathbb{R}^{(m'+n') \times D}$, where $D$ is the feature dimension.

\subsection{Dual-head Structure for Change Classification and Cross-visibility Confidence Estimation}\label{subsec:cls_head}
\begin{figure}[!t]
	\captionsetup{font=footnotesize}
	\centering
	\begin{subfigure}[b]{0.15\textwidth}
        \centering
        \includegraphics[width=1.0\textwidth]{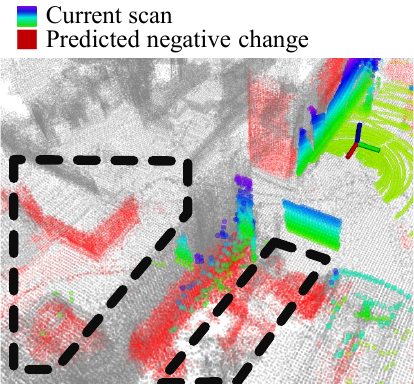}
        \caption{}
        \label{fig:conf_head_a}
    \end{subfigure}
    \begin{subfigure}[b]{0.15\textwidth}
        \centering
        \includegraphics[width=1.0\textwidth]{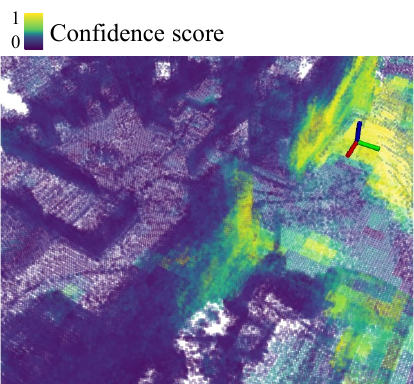}
        \caption{}
        \label{fig:conf_head_b}
    \end{subfigure}
    \begin{subfigure}[b]{0.15\textwidth}
        \centering
        \includegraphics[width=1.0\textwidth]{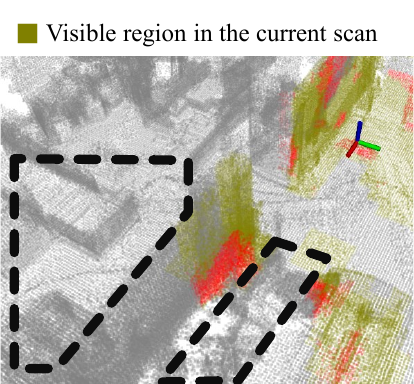}
        \caption{}
        \label{fig:conf_head_c}
    \end{subfigure}
	\caption{Our approach {handles} occlusion using cross-visibility confidence. 
    Gray points denote static regions, and red points indicate negative \changed{change} predictions by the class head. 
    (a) Occluded areas (black dashed) caused by walls hinder accurate prediction. 
    (b) The confidence head estimates visibility scores. 
    (c) Class predictions are applied only to visible regions, reducing errors in occluded areas.}
	\label{fig:conf_head_works}
	\vspace{-0.5cm}
\end{figure}

\resubmit{LiDAR map and scan data often exhibit different viewpoint coverages and suffer from occlusion, 
making direct geometric comparisons or na\"{\i}ve change classification unreliable for change detection, as shown in \figref{fig:conf_head_works}(a). 
To address this, we formulate change detection as a joint reasoning problem that estimates both the changed class and the cross-visibility confidence. 
As illustrated in \figref{fig:dual_head_architecture}, this formulation is implemented through a dual-head structure consisting of a class head and a confidence head.
The class head predicts whether each point is a change or static,
while the confidence head estimates the probability that each point is visible in both the map and scan.}
The class head is trained using a cross-entropy loss:
\vsfigu
\begin{equation}
  \mathcal{L}_\mathrm{cls} = -\frac{1}{m+n} \sum_{j=1}^{m+n} \sum_{c=0}^{2} y_{j,c} \log(\hat{y}_{j,c}),
  \vsfigu
\end{equation}
where $y_{j,c}$ and $\hat{y}_{j,c}$ denote the ground truth and predicted class probabilities for the $j$-th point, respectively.
Here, the class label $c \in \{0, 1, 2\}$ indicates static~(0), a positive change~(1), or a negative change~(2).

\resubmit{
  To complement the class head, the confidence head predicts a cross-visibility confidence score $c(d_{jk})$,
  which quantifies the likelihood that a point is jointly visible in both regions.
  High-confidence points are reliably observed in both, while low-confidence points are likely occluded in one. 
  The confidence score is computed based on the spatial distance to the nearest point in the opposite domain,
  under the assumption that larger distances imply higher occlusion probability.}
Let $p_j$ and $p_k$ denote the nearest points in the map and scan, and let $d_{jk}$ be their Euclidean distance.
We define the ground truth confidence $c(d_{jk})$ using a exponential decay function:
\begin{equation}
  c(d_{jk}) = 
  \begin{cases}
  	0, & \text{if } d_{jk} \geq \tau_\mathrm{ocl}, \\
  \exp\left(-\lambda \cdot (d_{jk} - {\tau_\mathrm{vox}})\right), & \text{if } {{\tau_\mathrm{vox}} < d_{jk} < \tau_\mathrm{ocl}}, \\
  1, & \text{if } {d_{jk} \leq {\tau_\mathrm{vox}}},
  \end{cases}
\label{eqn:confidence}
\end{equation}
where $\lambda$ represents the exponential decay rate, ${\tau_\mathrm{vox}}$ is the voxel size, and $\tau_\mathrm{ocl}$ is the occlusion distance threshold.
To improve training stability and convergence speed, we truncate the decay at $\tau_\mathrm{ocl}$ instead of applying it across the full range of $d_{jk}$.
The confidence head is trained using mean squared error (MSE) loss:
\begin{equation}
  \mathcal{L}_\mathrm{conf} = \frac{1}{|\mathcal{P}|} \sum_{(j,k) \in \mathcal{P}} \left\| c(d_{jk}) - \hat{c}_j \right\|^2,
  \vsequ
\end{equation}
where $\mathcal{P}$ is a set of randomly sampled point pairs $(p_j, p_k)$ from the map and scan, with {$|\mathcal{P}|$} denoting the number of sampled pairs. 
{$\hat{c}_j$} is the predicted confidence value from the confidence head.
During training, the class and confidence heads are provided with distinct input features tailored to their respective objectives.
Since change classification requires high-level semantic understanding, we feed the final backbone feature map $F_\mathrm{HLF}$ as input to the class head. 
In contrast, cross-visibility estimation relies more on local geometric cues, so we use earlier-stage features $F_\mathrm{LLF}$ for the confidence head.
The effectiveness of this backbone feature division will be explored in {more} detail {in} {\secref{subsec:exp_ablation}}.

The total training loss is the weighted sum of the classification loss and the confidence loss:
\begin{equation}
  \mathcal{L} = \mathcal{L}_\mathrm{cls} + \alpha \mathcal{L}_\mathrm{conf},
  \vsfigu
\end{equation}
where the confidence loss weight $\alpha$ balances the contributions of the main task of change classification and the auxiliary supervision of confidence scores.

\subsection{Probabilistic Change Integration and Prior Map Update}\label{subsec:prob_map_update}
\resubmit{Our method yields $N$ change detection results on the same prior map for $N$ current observations. 
For reliable inference, both scan and map predictions are conditioned on cross-visibility confidence. 
$\tau_{\mathrm{conf}}^{\mathrm{scan}}$ and $\tau_{\mathrm{conf}}^{\mathrm{map}}$ are the confidence thresholds for scan and map, used to suppress false changes and filter unreliable regions, respectively. 
Scan regions predicted as positive changes with $\hat{c}_j < \tau_{\mathrm{conf}}^{\mathrm{scan}}$ are classified as \textit{changes}, 
while map updates use a recursive Bayesian filter that integrates observations over time and excludes low-visibility regions prone to occlusion errors. 
The update process is defined as follows:}
\vspace{-0.05cm}
\begin{equation}
  \begin{aligned}
  &l(\hat{y}_{j} \mid \mathcal{M}_{0:t}) \\
  &=
  \left\{
    \begin{array}{ll}
      \begin{aligned}[t]
      &l(\hat{y}_{j} \mid \mathcal{M}_{0:t-1}) \\
      &+ l(\hat{y}_{j} \mid \mathcal{M}_{t}) - l(\hat{y}_{j}),
      \end{aligned}
      & \vcenter{\hbox{\(\text{if } \hat{c}_j > \tau_{\mathrm{conf}}^{\mathrm{map}},\)}} \\
      l(\hat{y}_{j}), 
      & \text{otherwise}
    \end{array}
  \right.
  \end{aligned}
\end{equation}
where $l(\hat{y}_j \mid \mathcal{M}_t)$ is the log-odds derived from the observation at time $t$, $l(\hat{y}_j)$ is the prior class probability, 
\changed{$\hat{c}_j$} is the \changed{predicted} confidence score.

\section{Experiments}\label{sec:experiments}

\subsection{Experimental Setup}
\label{subsec:exp_setup}

The purpose of our experiments was to evaluate our method's performance in change detection and its effectiveness in updating the prior map.


\textbf{Datasets.} We used two datasets for quantitative evaluation. 
  The first is our custom dataset, which consists of a prior map built in August {2024} and 1,000 scan frames collected in October {2024} 
  \resubmit{across} three \resubmit{real-world} environments, including construction sites and a laboratory. 
  \resubmit{These environments contain semantically diverse objects (e.g., ladders, rubber cones, and various experimental structures) and exhibit substantial environmental variability.} The ground-truth change labels were manually annotated.
  To assess the generalization {capability} of our approach, we also used the LiSTA dataset~\cite{joseph2024icra}, a changing indoor office dataset, 
  also annotated with ground truth change labels.

\textbf{Comparison methods.} We compared our method with both classical and learning-based baselines: {v}isibility-based, occupancy-based, MapMOS, and SPS. 
\changed{The visibility}-based method~\cite{underwood2013icra, kim2022icra} detects changes based on differences in visibility between the map and scan.
\changed{The occupancy}-based method~\cite{walcot2012iros} considers points in the same voxel as static if both map and scan points are present, 
and as change if only map or scan points are present. 
MapMOS~\cite{mersch2023ral} identifies high-dynamic objects through map-scan comparison and volumetric fusion; we retrained it on our pseudo-labeled data, 
as the original model was trained solely on the SemanticKITTI-MOS dataset.
SPS~\cite{hroob2024ral} predicts a stability score via a self-supervised network, with points below a fixed threshold are classified as changes.

\textbf{Evaluation metric.} We evaluate the performance of \changed{each} method in both scan-wise and map-wise change detection.
Scan-wise positive change detection is evaluated using the intersection over union~(IoU), defined as:

\newcommand{\iou}{\text{IoU}_{\text{PC}}}
\newcommand{\PRmap}{\text{PR}}
\newcommand{\RRmap}{\text{RR}}

\begin{equation}
  \changed{\iou=\frac{\mathrm{TC}}{\mathrm{TC}+\mathrm{FC}+\mathrm{FS}}},
\label{eqn:iou}
\end{equation}

\noindent \changed{where $\mathrm{TC}$, $\mathrm{FC}$, and $\mathrm{FS}$ denote the true changes, false changes, and false statics}, respectively.
For map-wise evaluation of negative change detection, we use the preservation rate (PR) and rejection rate (RR), following the metrics modified from~\cite{lim2021ral}:

\begin{itemize}
	 \setlength{\itemsep}{0.5em}
	\item $\PRmap = \textcolor{black}{\frac{\text{\# of preserved static points}}{\text{\# of total static points on the prior map}}}$,
	
  \item $\RRmap = \textcolor{black}{1- \frac{\text{\# of remaining negative changes}}{\text{\# of total negative changes on the prior map}}}$,
	
  \item $\text{F}_1 = 2\PRmap \cdot \RRmap / (\PRmap + \RRmap)$.

\end{itemize}

We adopt different metrics for evaluating the map and scans, since the scans consist of multiple frames, 
whereas the map is a single, accumulated dataset.
Applying identical metrics would be inappropriate due to this structural difference.  

\textbf{Implementation details.}
\changed{For training, we augment a custom dataset, collected from a different environment than the evaluation set, 
using the method in~\secref{subsec:mono_temporal_datagen}, with a voxel size of 0.1m used for input quantization}.
The confidence loss weight $\alpha$ is set to 0.01.
For confidence score calculation (see~\eqref{eqn:confidence}), we set $\tau_{\mathrm{ocl}} = 3.0\,\mathrm{m}$ and $\lambda = 10$. 
Training is performed for up to 50 epochs with a batch size of 2 using the Adam optimizer~\cite{Kingma2014AdamAM}, 
and the best validation model is selected for evaluation. 
During inference, dataset-specific optimal confidence thresholds $\tau_{\mathrm{conf}}^{\mathrm{map}}$ and $\tau_{\mathrm{conf}}^{\mathrm{scan}}$  are applied for the Custom and LiSTA datasets 
to ensure reliable scan and map updates, as discussed in~\secref{subsec:exp_ablation}.
\resubmit{Before evaluating the LD change detection performance, high-dynamic~(HD) objects are removed from both the map and scan.
The effect of removing HD objects is analyzed in~\secref{subsec:exp_hd_removal}}.

\begin{table}[!t]
    \centering
    \captionsetup{font=footnotesize}
    \caption{Change detection performance comparison on the custom dataset {in terms of scan-wise} IoU and map-wise PR, RR, and $\mathrm{F_1}$ scores. Best results in \textbf{bold}, second best in \hl{a gray background}.}
    \setlength{\tabcolsep}{7pt}
    {\scriptsize
        \label{tab:custom_evaluation}
        \begin{tabular}{llcccc}
        \toprule
        \multirow{2}{*}[-0.5em]{Seq.} & \multirow{2}{*}[-0.5em]{Method}  & \multicolumn{1}{c}{Scan-wise} & \multicolumn{3}{c}{Map-wise} \\ 
        \cmidrule(lr){3-3} \cmidrule(lr){4-6} 
                               &   & IoU{~$\uparrow$}       & PR{~$\uparrow$}               & RR{~$\uparrow$}              & $\mathrm{F_1}${~$\uparrow$} \\  \midrule
        
        \multirow{4}{*}{{\texttt{\changed{Const}-1F}}}      & Visibility~\cite{underwood2013icra}              & 0.069          & 0.671            & 0.969           & \textbf{0.793}  \\
                                                            & Occupancy~\cite{walcot2012iros}                  & 0.297          & 0.629            & 0.863           & {0.728}      \\
                                                            & MapMOS~\cite{mersch2023ral}                      & \hl{0.679}        & 0.984            & 0.182           & 0.308           \\
                                                            & SPS~\cite{hroob2024ral}                          & 0.466          & 0.035            & {0.999}         & 0.069           \\
                                                            & {Chamelion~(ours)}                               & \textbf{0.728}        & 0.950            & 0.616           & \hl{0.747}      \\ \midrule

        \multirow{4}{*}{{\texttt{\changed{Const}-2F}}}      & Visibility~\cite{underwood2013icra}              & 0.013          & 0.705            & 0.915           & \hl{0.796}           \\
                                                            & Occupancy~\cite{walcot2012iros}                  & 0.486          & 0.635            & 0.874           & 0.736      \\
                                                            & MapMOS~\cite{mersch2023ral}                      & \hl{0.480}     & {0.998}          & 0.000           & 0.000           \\
                                                            & SPS~\cite{hroob2024ral}                          & 0.439          & 0.125            & {0.999}         & 0.222           \\
                                                            & {Chamelion~(ours)}                               & \textbf{0.484} & {0.954}          & {0.839}         & \textbf{0.893}  \\ \midrule

        \multirow{4}{*}{{\texttt{Lab}}}                     & Visibility~\cite{underwood2013icra}              & 0.127          & 0.695            & {0.908}         & \hl{0.787}           \\
                                                            & Occupancy~\cite{walcot2012iros}                  & 0.259          & 0.651            & 0.871           & {0.745}           \\
                                                            & MapMOS~\cite{mersch2023ral}                      & \hl{0.757}     & {0.992}          & 0.166           & 0.284           \\
                                                            & SPS~\cite{hroob2024ral}                          & 0.394          & 0.023            & {0.998}         & 0.046           \\
                                                            & {Chamelion~(ours)}                               & \textbf{0.767} & {0.955}          & 0.713           & \textbf{0.816}  \\ \bottomrule
        \end{tabular}
    }
    \vsfig
\end{table}
\begin{figure*}[!t]
    \centering
    \captionsetup{font=footnotesize}
    \renewcommand{\arraystretch}{0} 
    \setlength{\tabcolsep}{2pt} 
    \begin{tabular}{ccccccc}
    \vsfiguu
        \begin{subfigure}{0.13\textwidth}
            \centering
            \includegraphics[width=\linewidth]{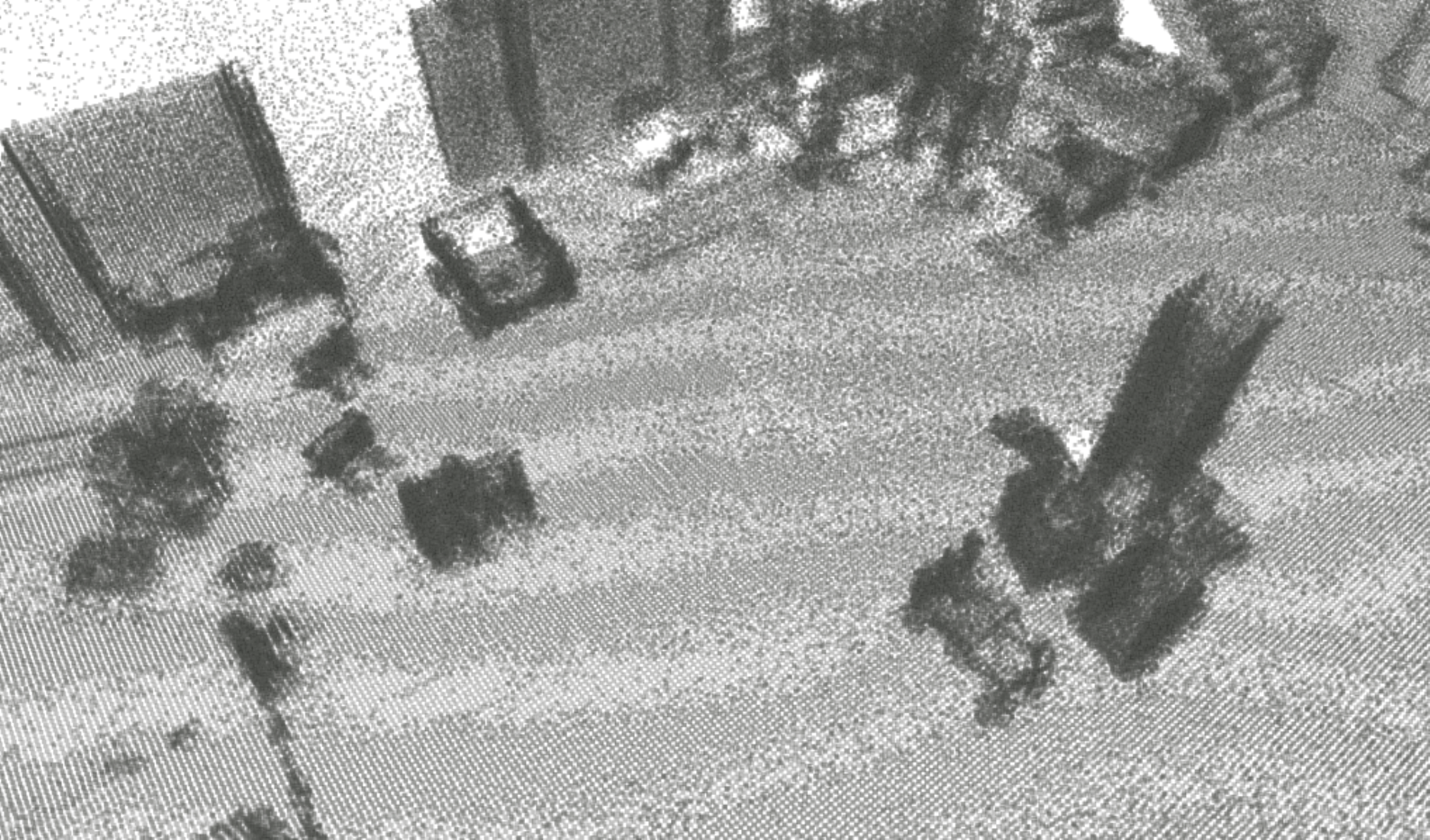}
        \end{subfigure} &
        \begin{subfigure}{0.13\textwidth}
            \centering
            \includegraphics[width=\linewidth]{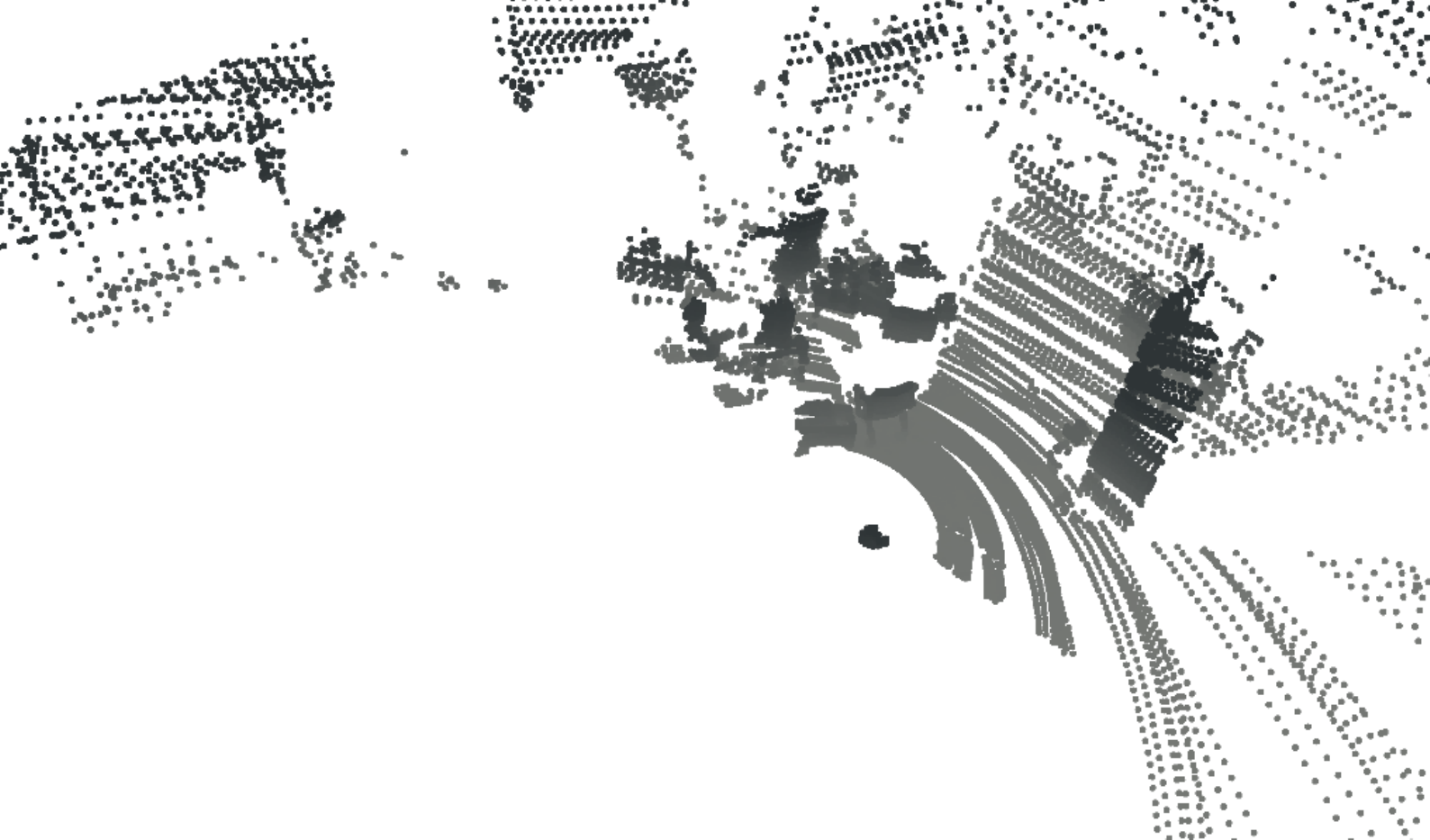}
        \end{subfigure} &
        \begin{subfigure}{0.13\textwidth}
            \centering
            \includegraphics[width=\linewidth]{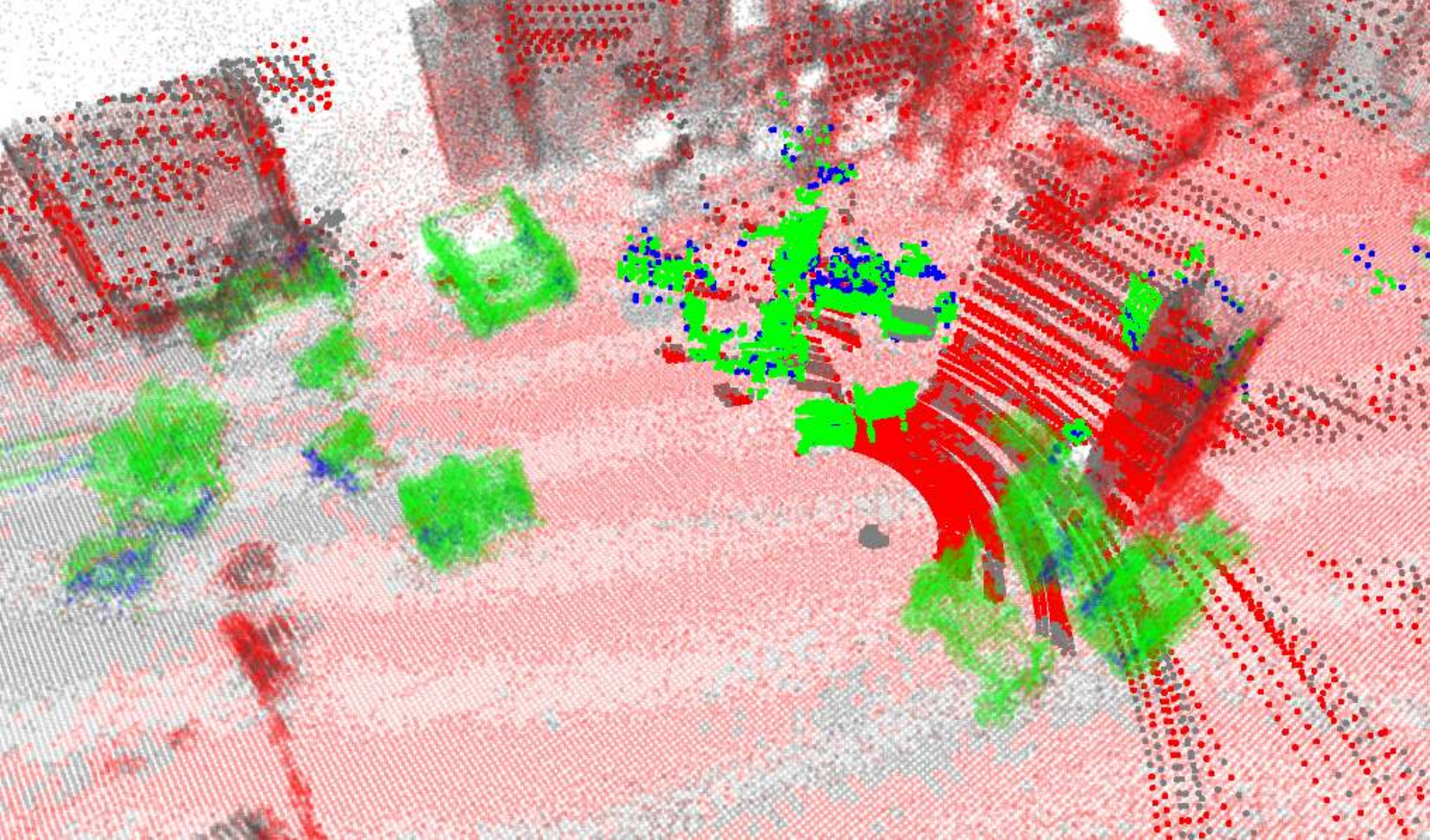}
        \end{subfigure} &
        \begin{subfigure}{0.13\textwidth}
            \centering
            \includegraphics[width=\linewidth]{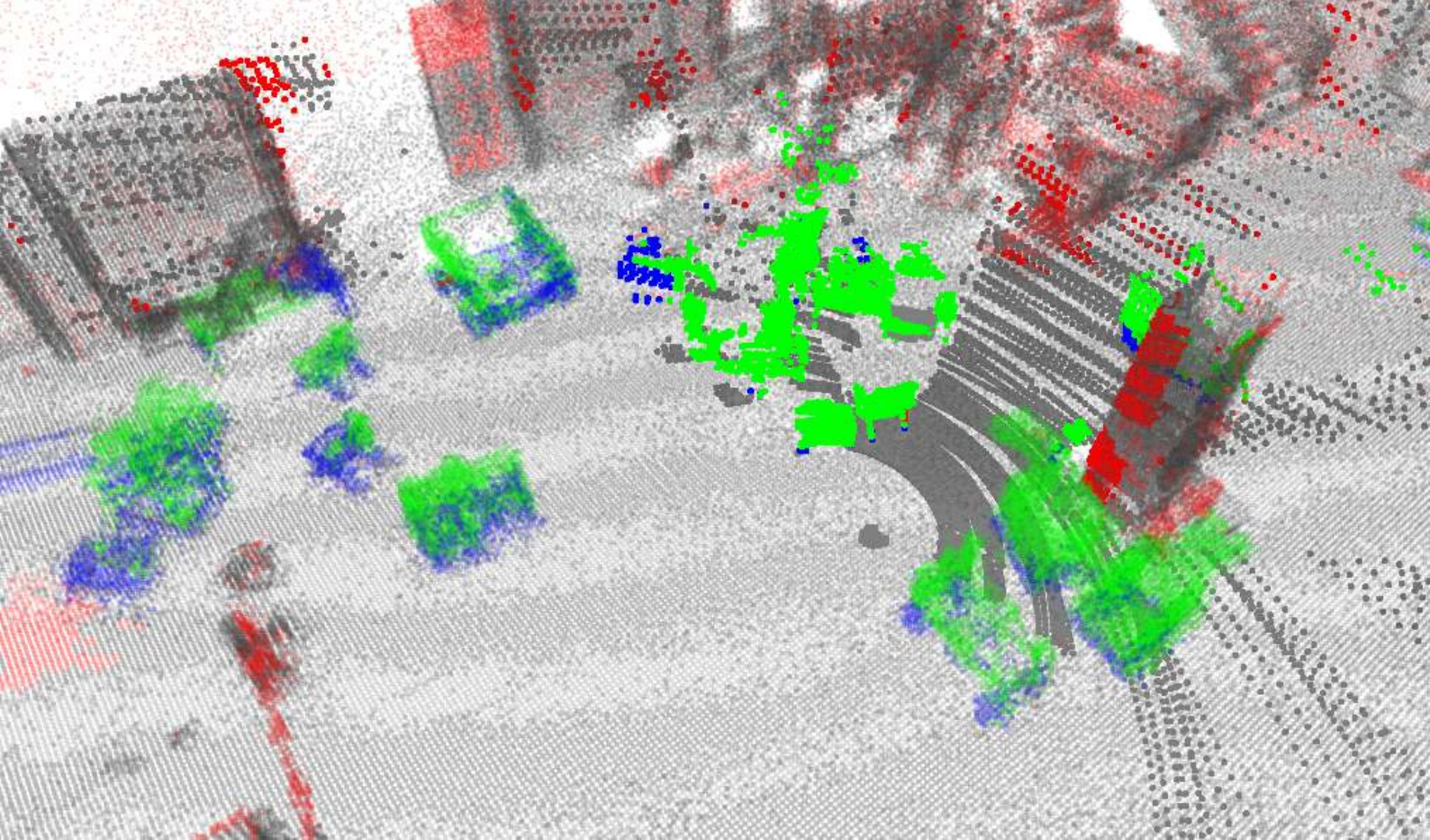}
        \end{subfigure} &
        \begin{subfigure}{0.13\textwidth}
            \centering
            \includegraphics[width=\linewidth]{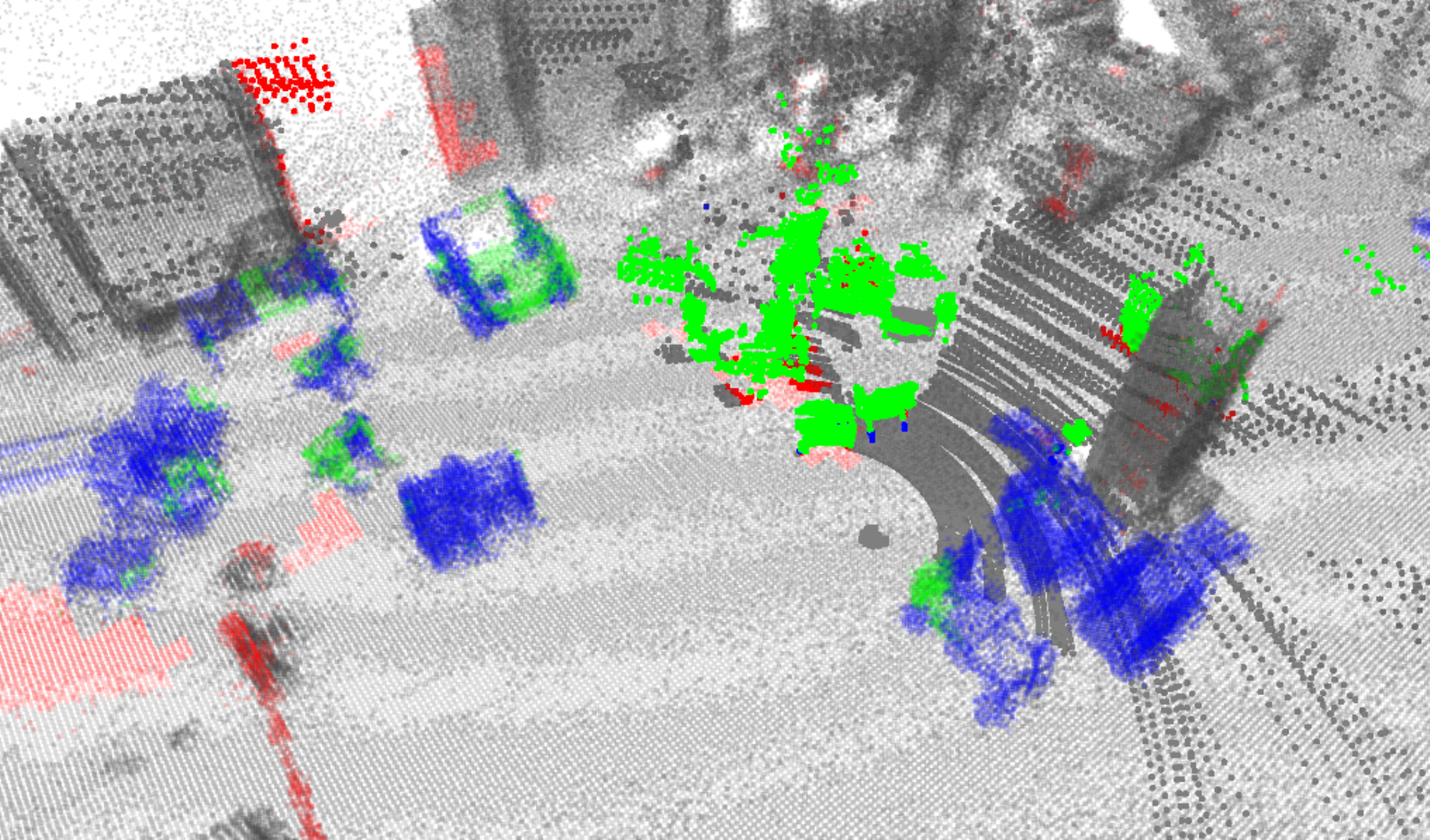}
        \end{subfigure} &
        \begin{subfigure}{0.13\textwidth}
            \centering
            \includegraphics[width=\linewidth]{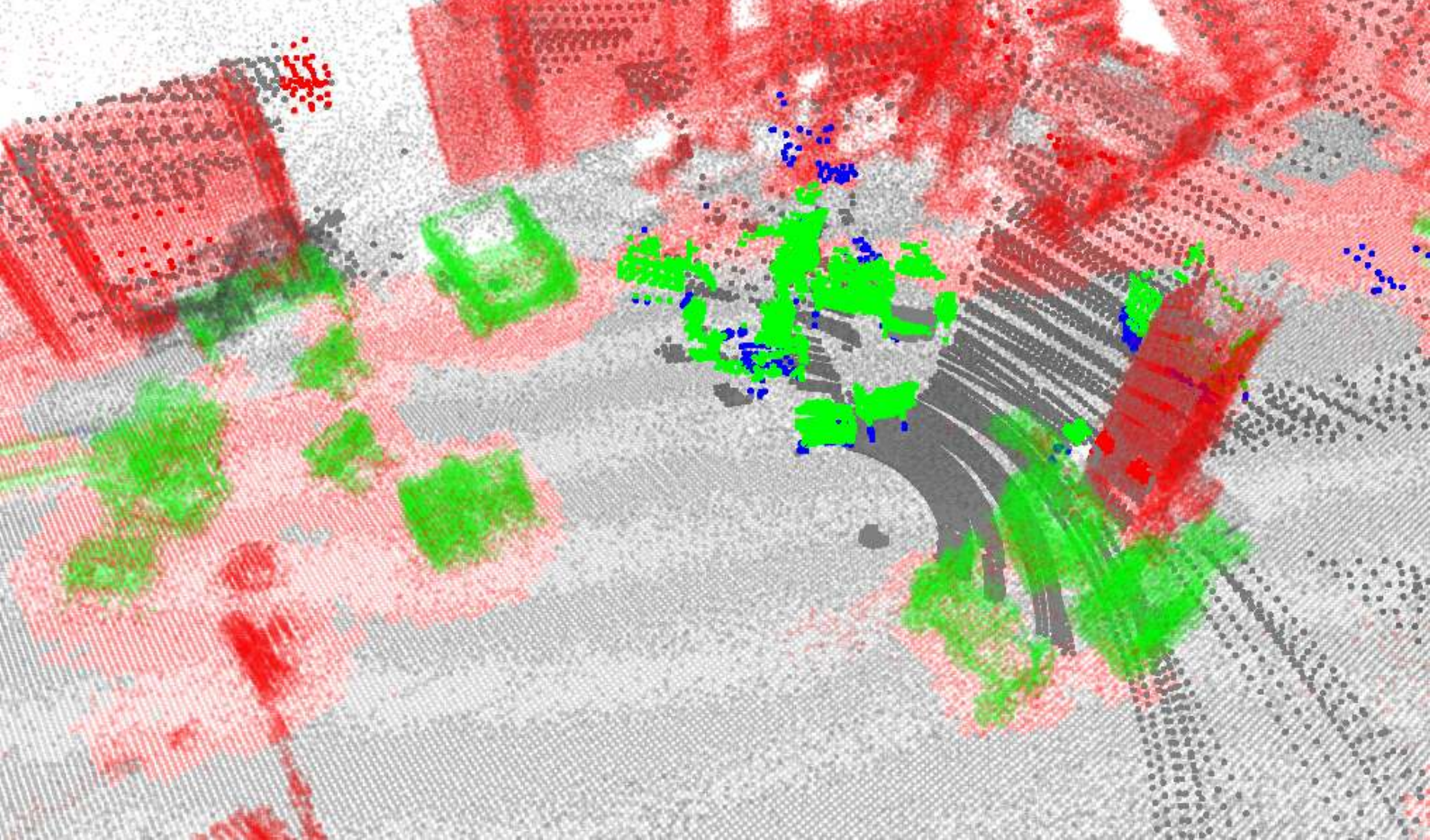}
        \end{subfigure} &
        \begin{subfigure}{0.13\textwidth}
            \centering
            \includegraphics[width=\linewidth]{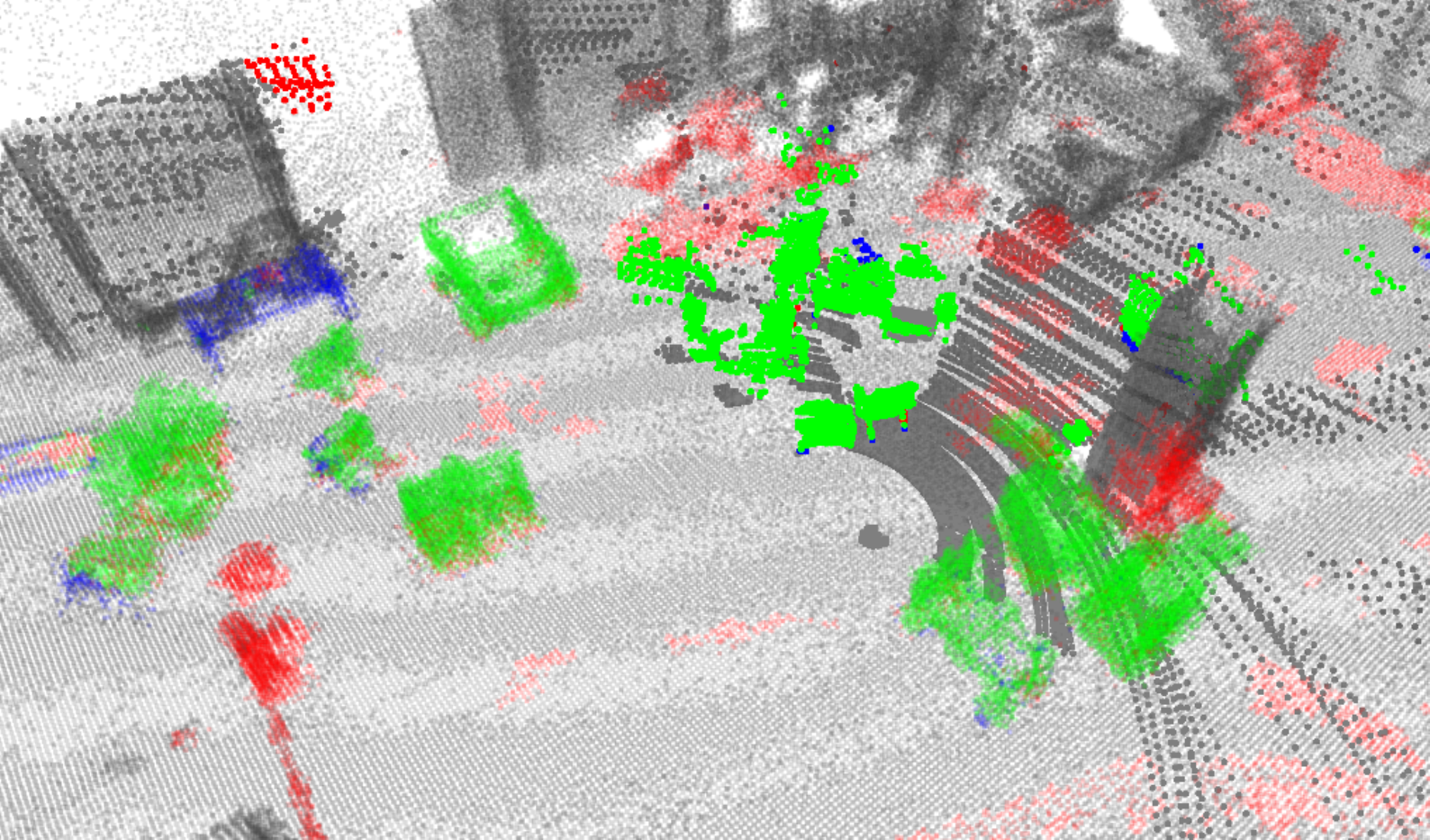}
        \end{subfigure} \\
        
        \begin{subfigure}{0.13\textwidth}
            \centering
            \includegraphics[width=\linewidth]{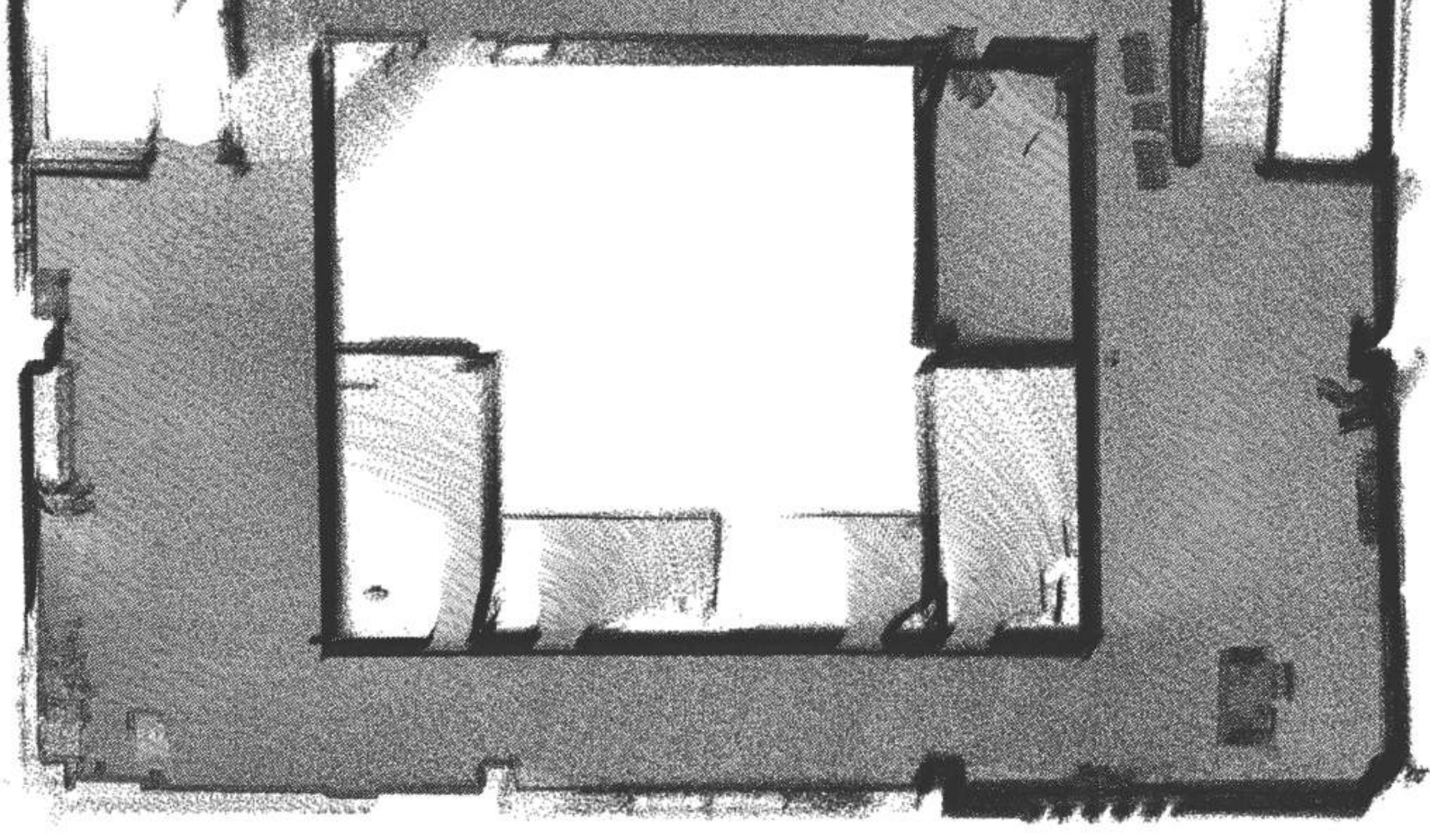}
        \end{subfigure} &
        \begin{subfigure}{0.13\textwidth}
            \centering
            \includegraphics[width=\linewidth]{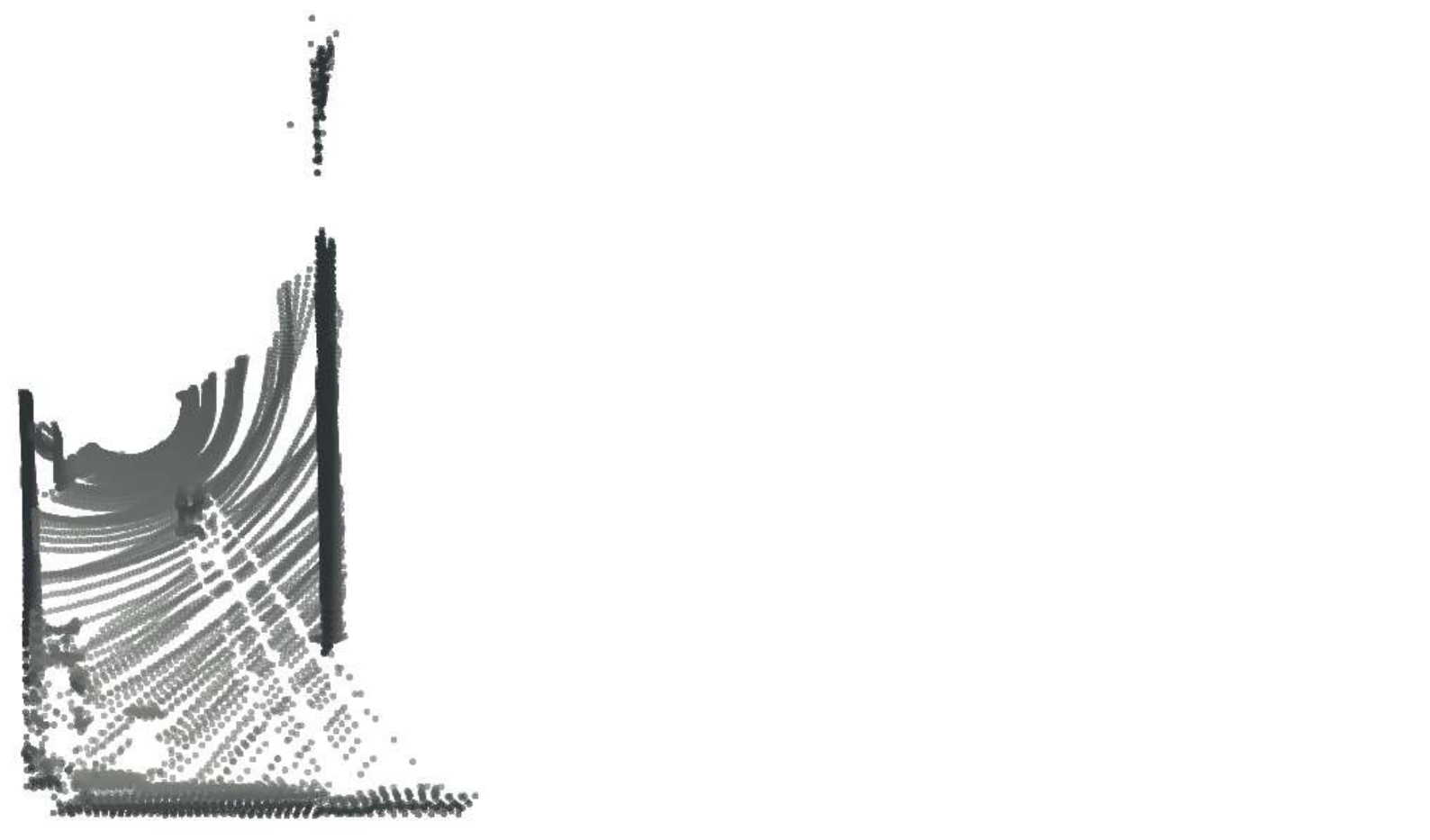}
        \end{subfigure} &
        \begin{subfigure}{0.13\textwidth}
            \centering
            \includegraphics[width=\linewidth]{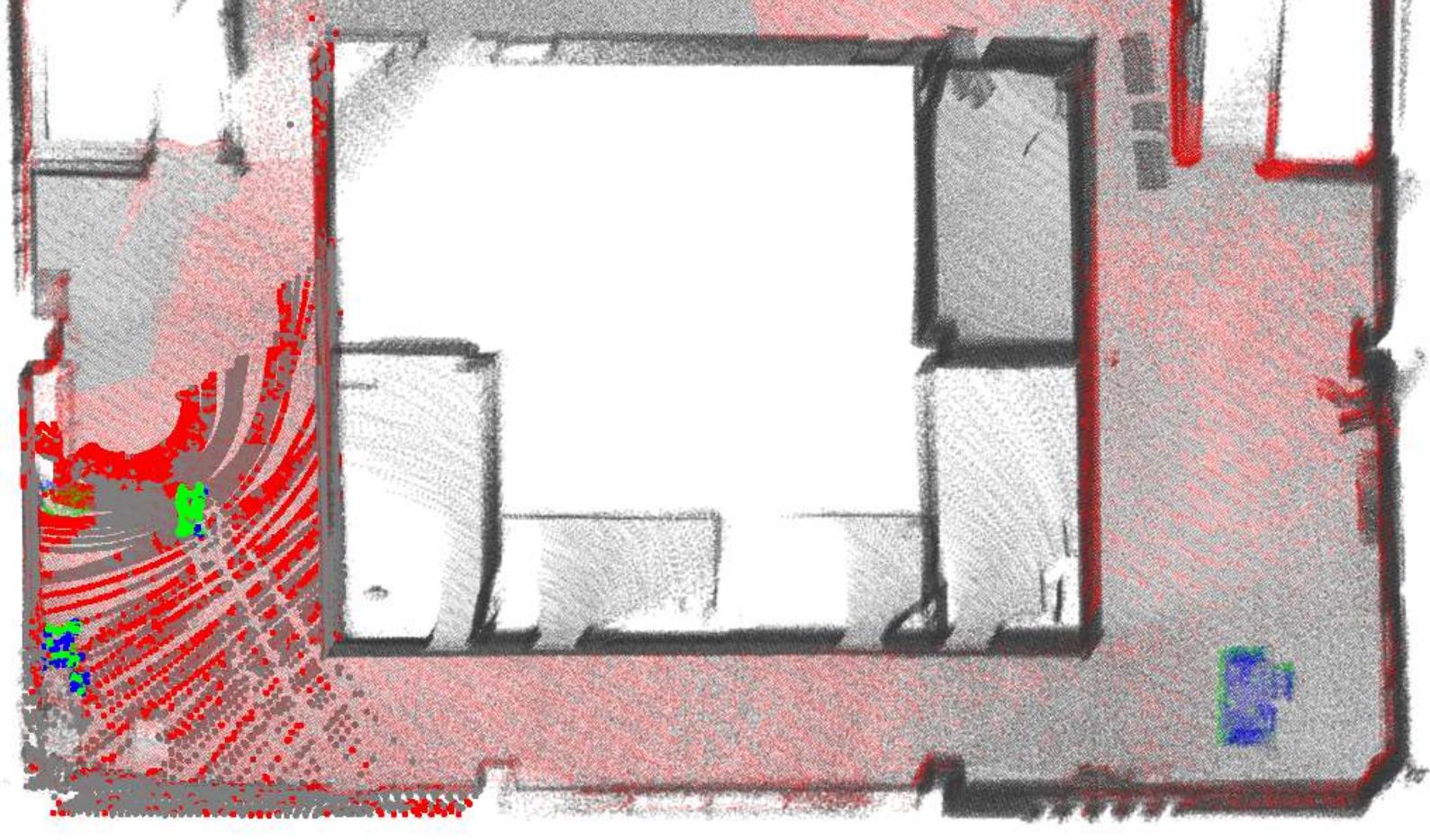}
        \end{subfigure} &
        \begin{subfigure}{0.13\textwidth}
            \centering
            \includegraphics[width=\linewidth]{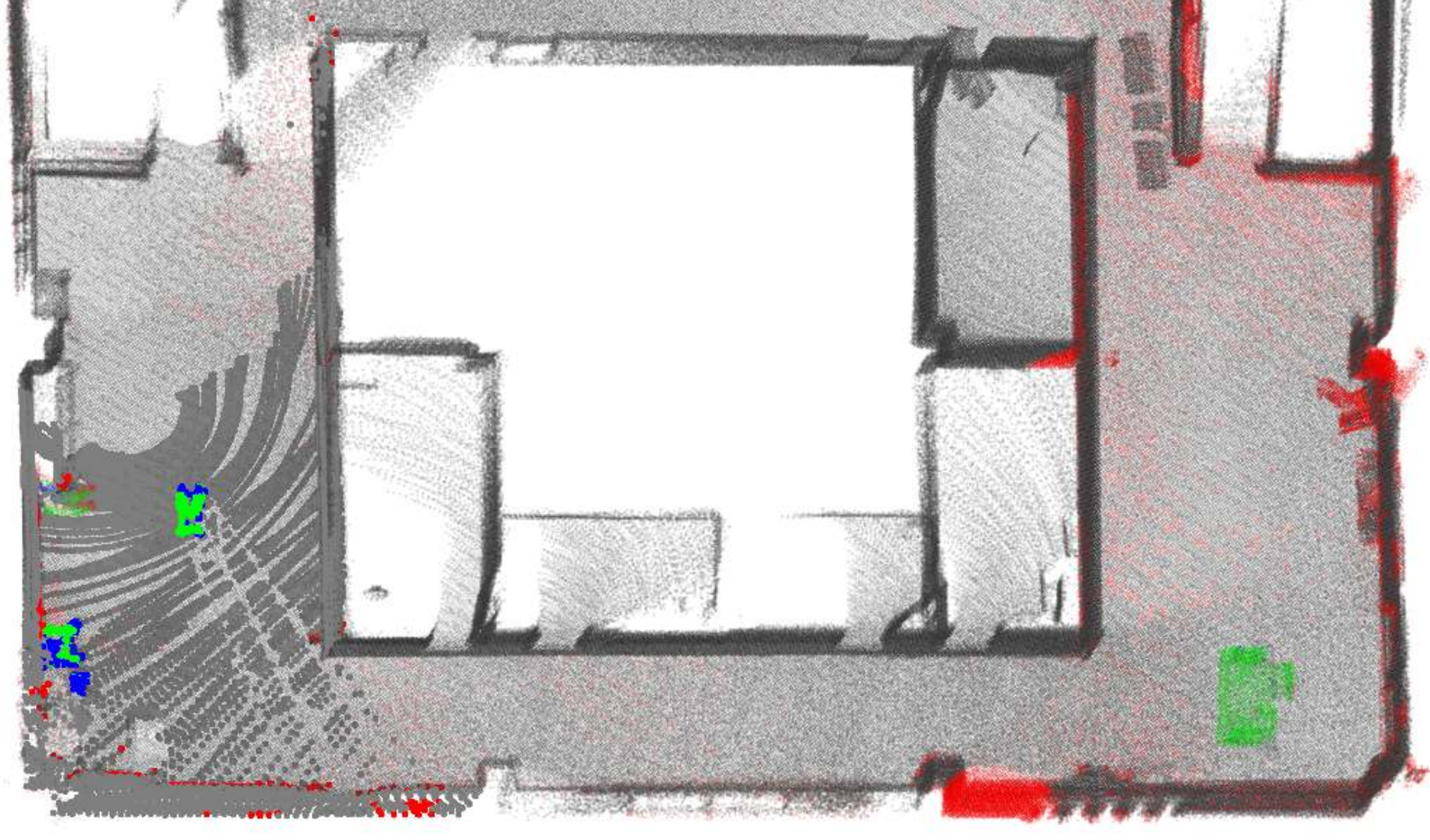}
        \end{subfigure} &
        \begin{subfigure}{0.13\textwidth}
            \centering
            \includegraphics[width=\linewidth]{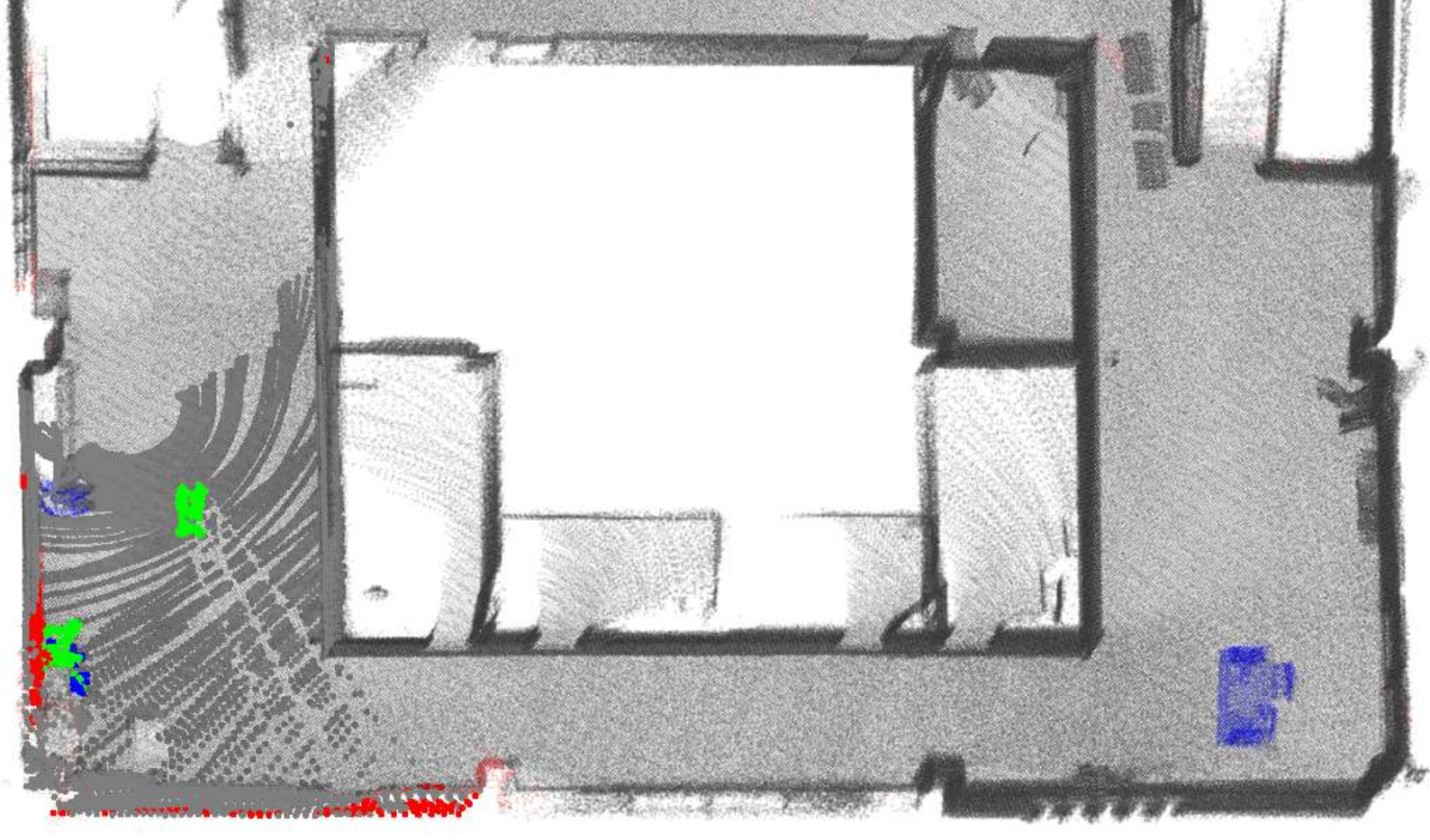}
        \end{subfigure} &
        \begin{subfigure}{0.13\textwidth}
            \centering
            \includegraphics[width=\linewidth]{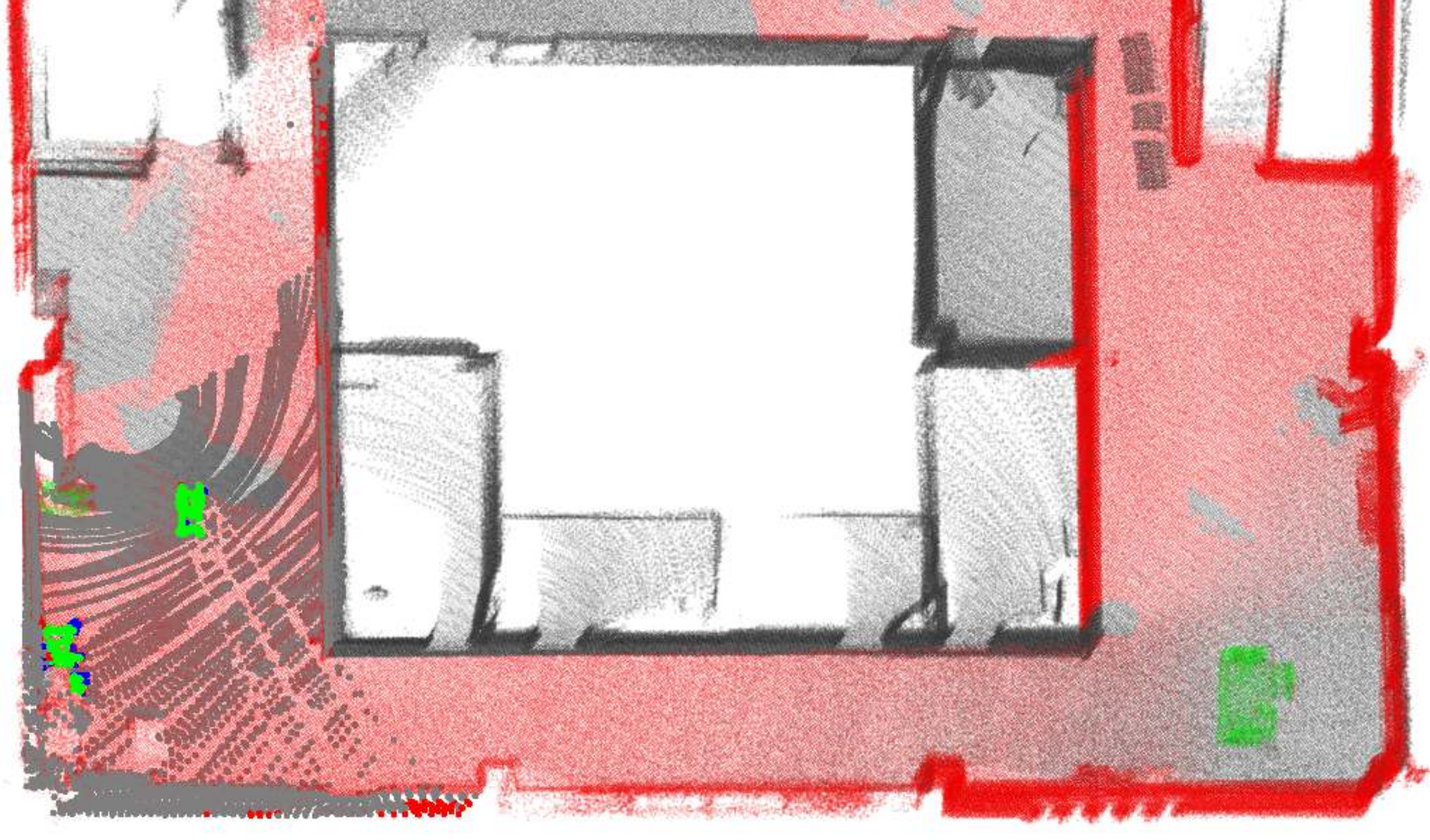}
        \end{subfigure} &
        \begin{subfigure}{0.13\textwidth}
            \centering
            \includegraphics[width=\linewidth]{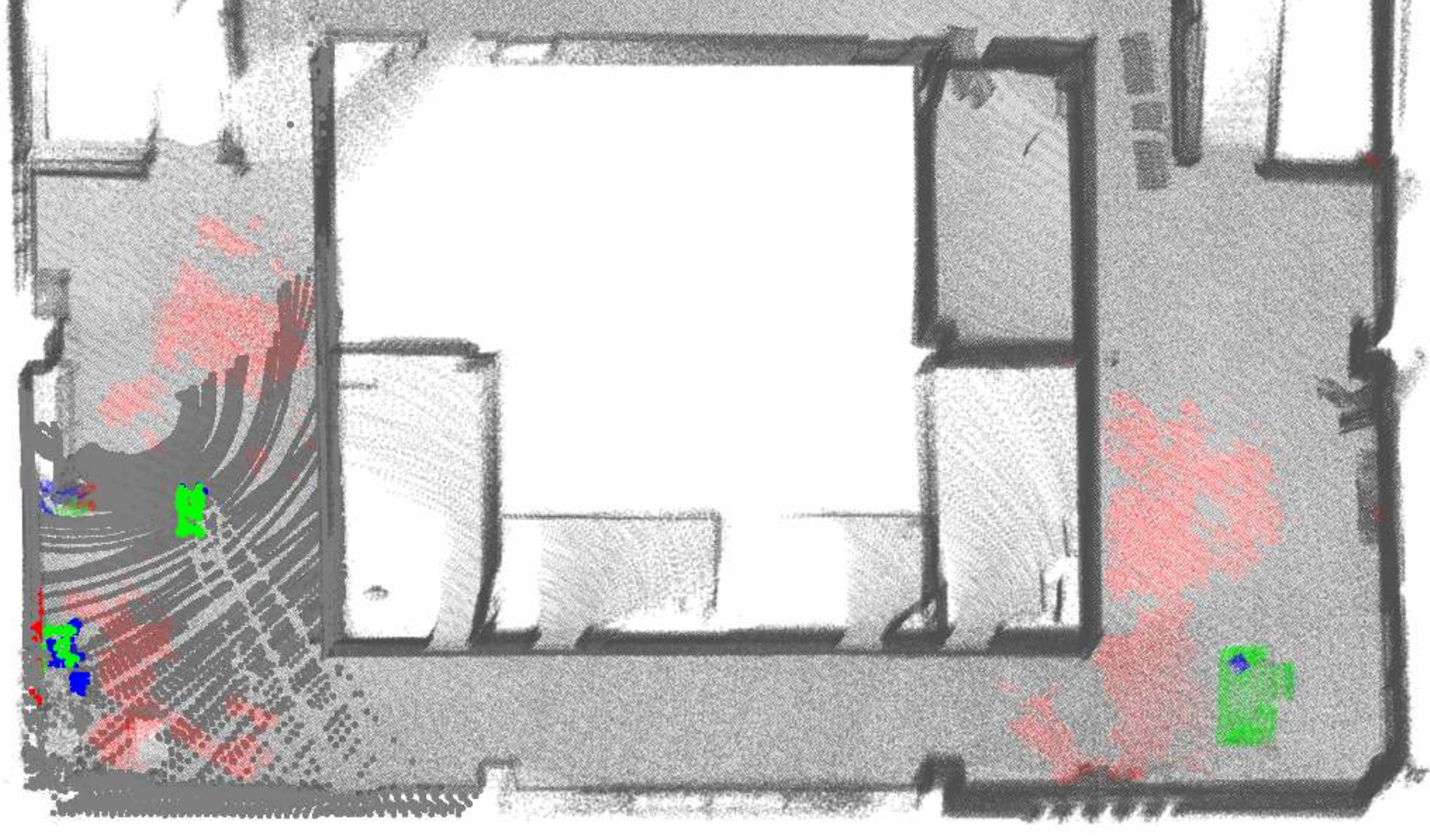}
        \end{subfigure} \\
        \vsfiguu
        \begin{subfigure}{0.13\textwidth}
            \centering
            \includegraphics[width=\linewidth]{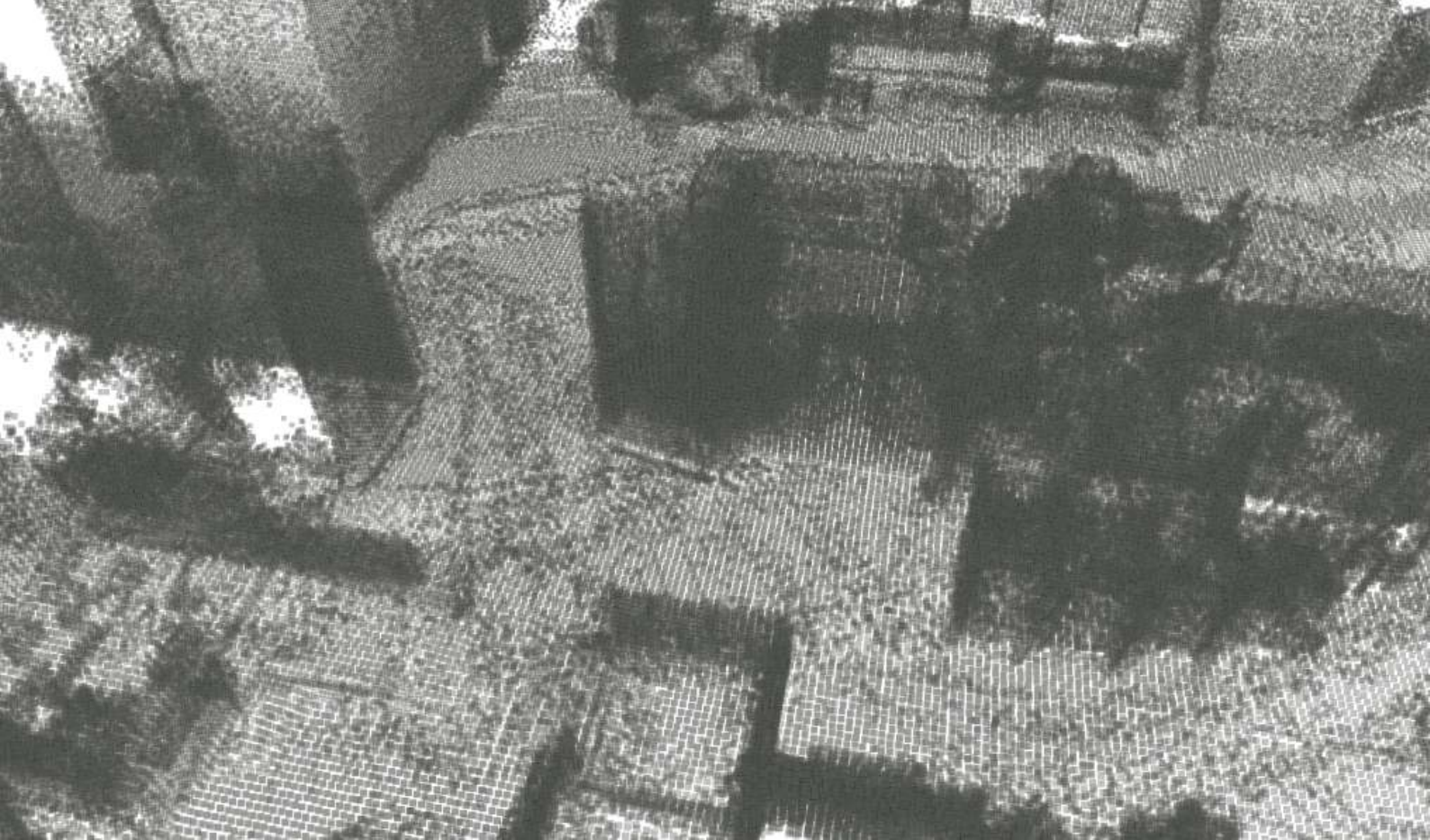}
            \caption{Prior map}
        \end{subfigure} &
        \begin{subfigure}{0.13\textwidth}
            \centering
            \includegraphics[width=\linewidth]{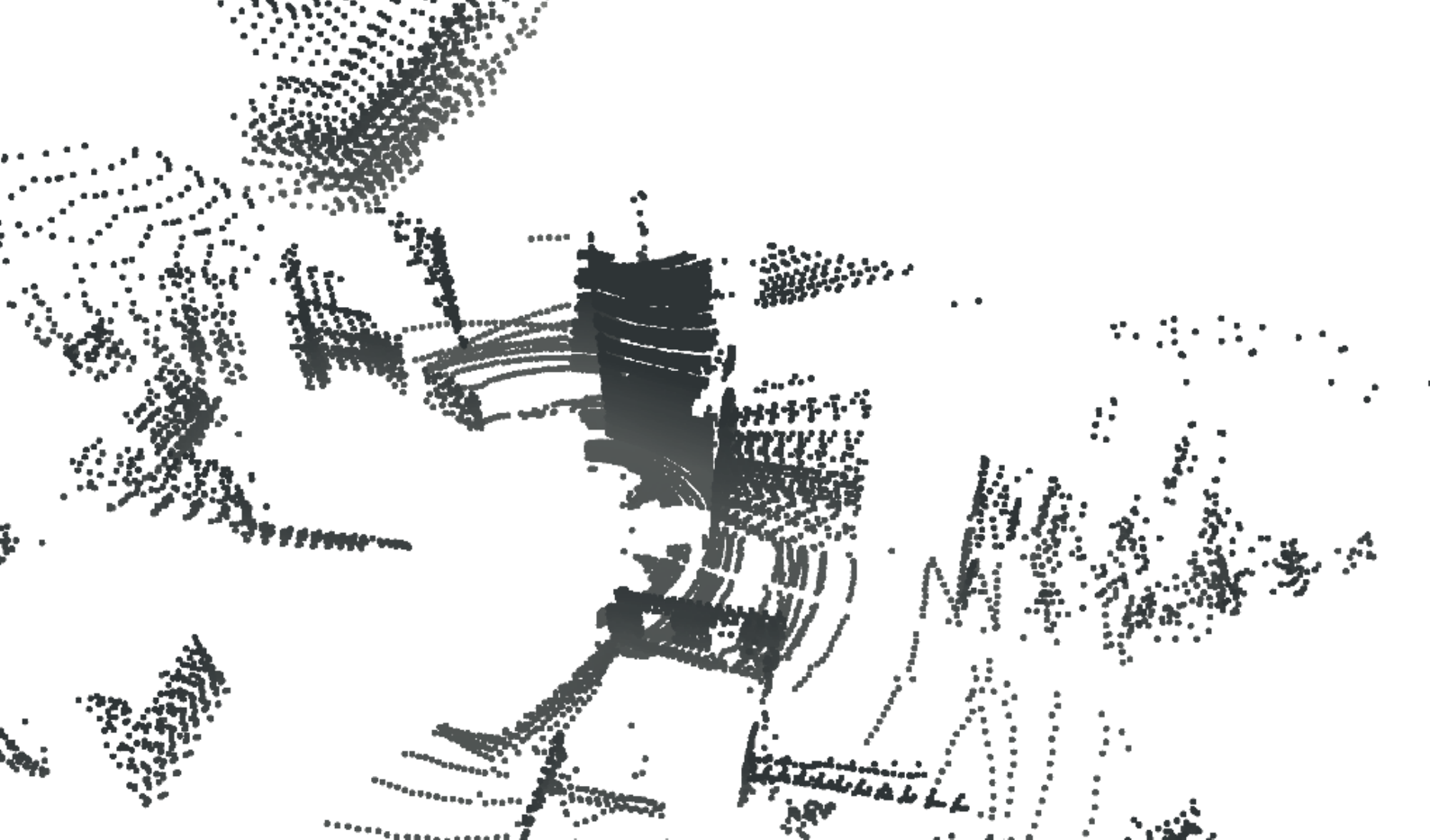}
            \caption{Current scan}
        \end{subfigure} &
        \begin{subfigure}{0.13\textwidth}
            \centering
            \includegraphics[width=\linewidth]{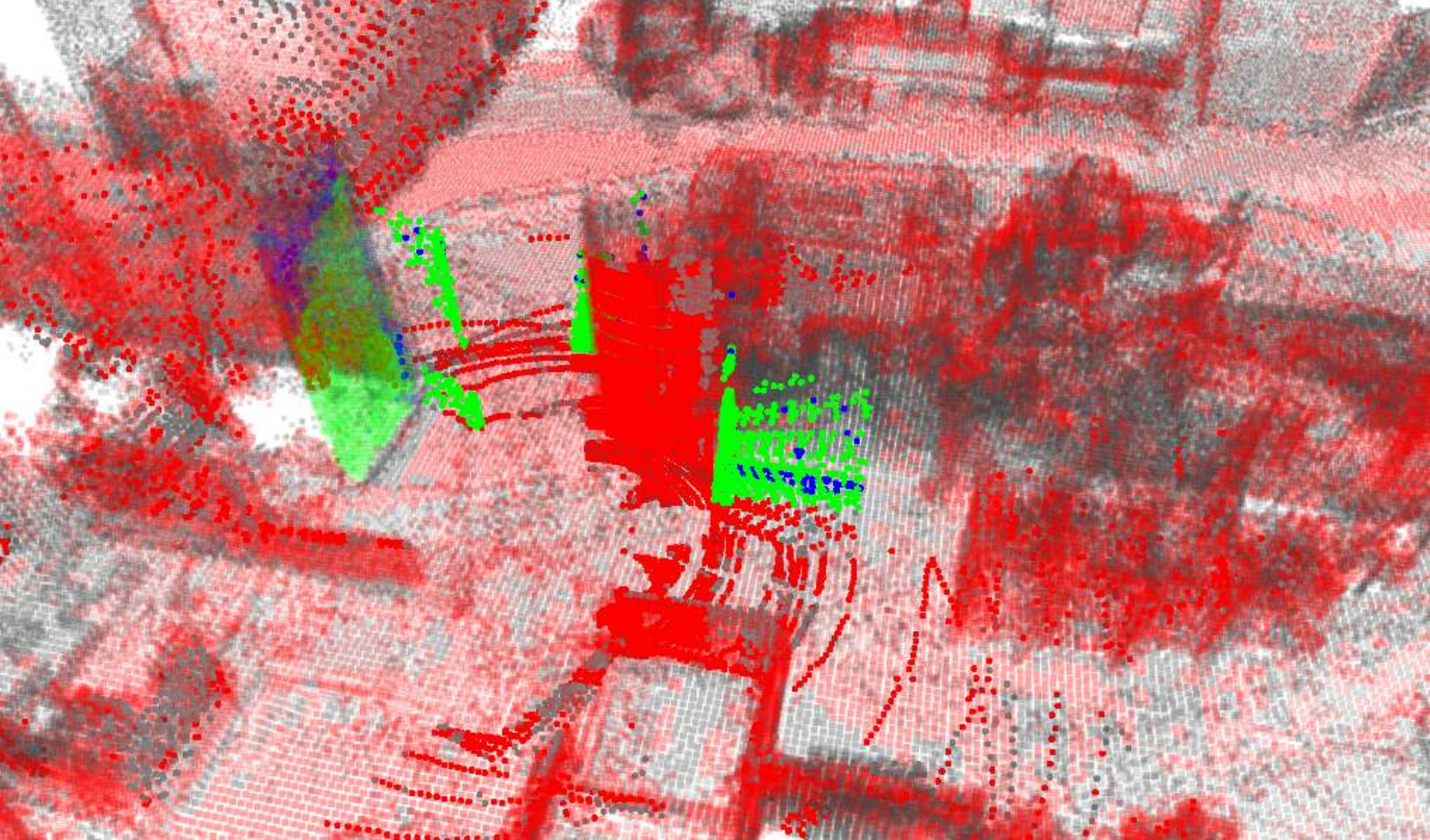}
            \caption{Visibility~\cite{underwood2013icra}}
        \end{subfigure} &
        \begin{subfigure}{0.13\textwidth}
            \centering
            \includegraphics[width=\linewidth]{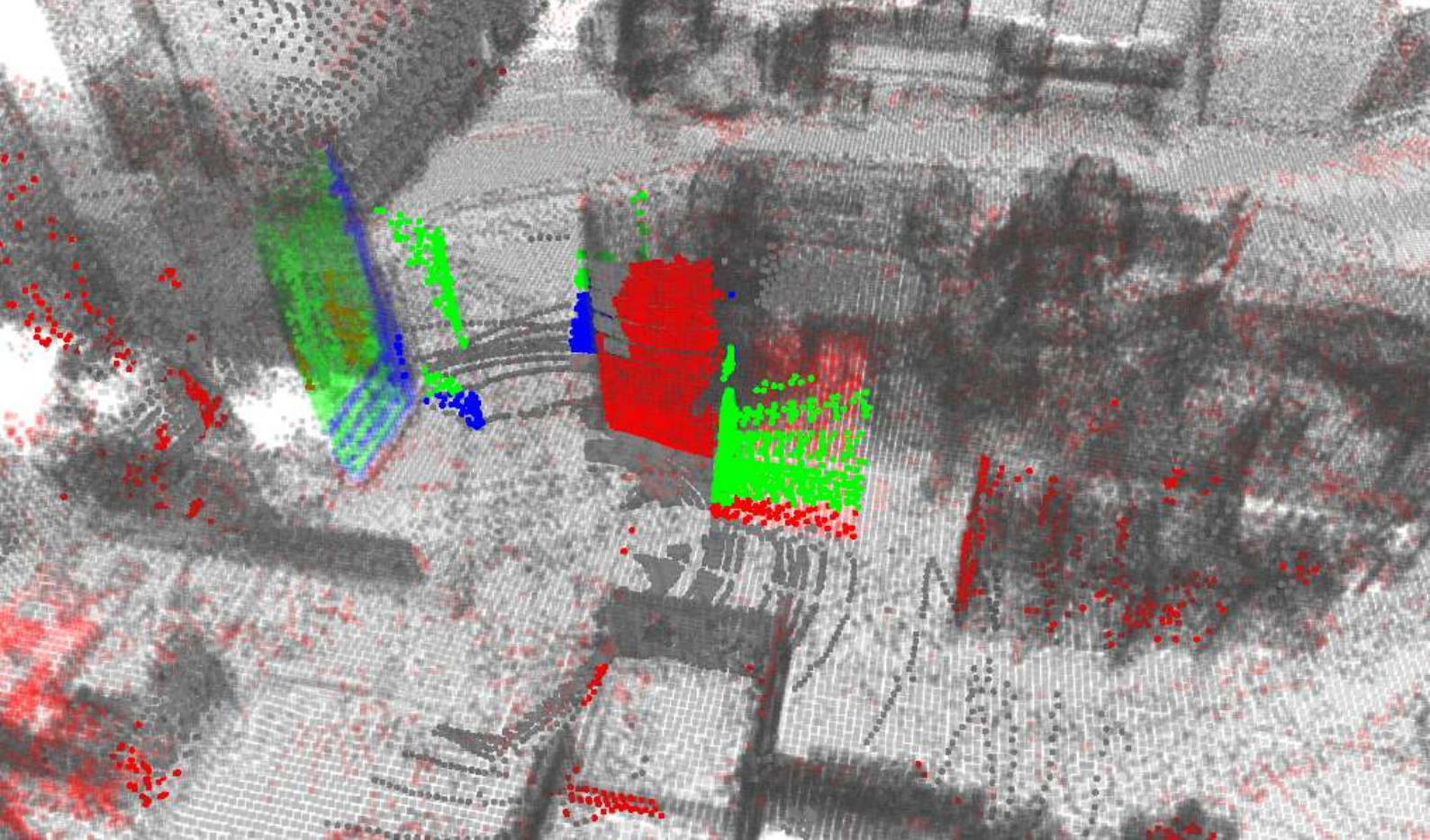}
            \caption{Occupancy~\cite{walcot2012iros}}
        \end{subfigure} &
        \begin{subfigure}{0.13\textwidth}
            \centering
            \includegraphics[width=\linewidth]{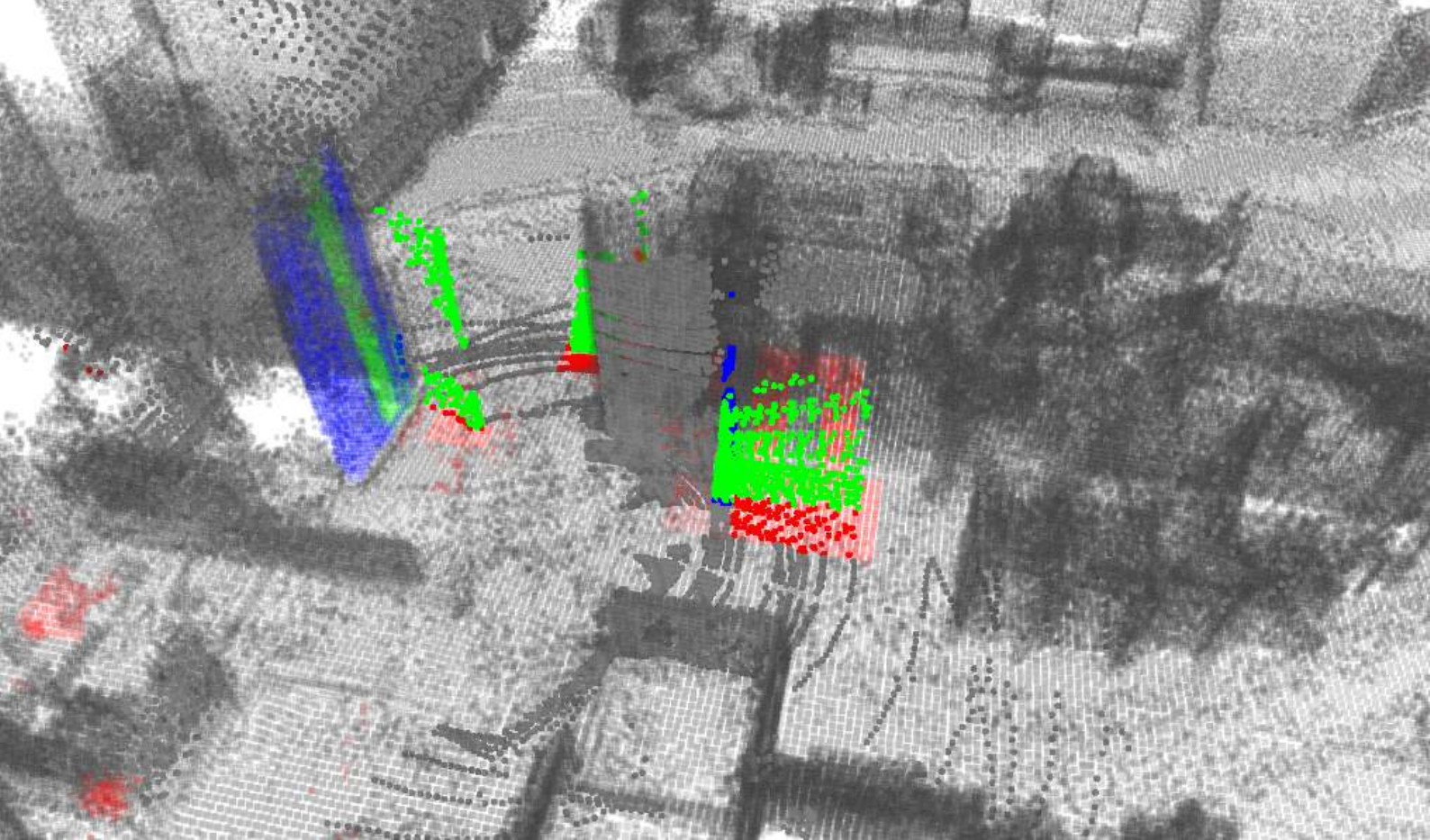}
            \caption{MapMOS~\cite{mersch2023ral}}
        \end{subfigure} &
        \begin{subfigure}{0.13\textwidth}
            \centering
            \includegraphics[width=\linewidth]{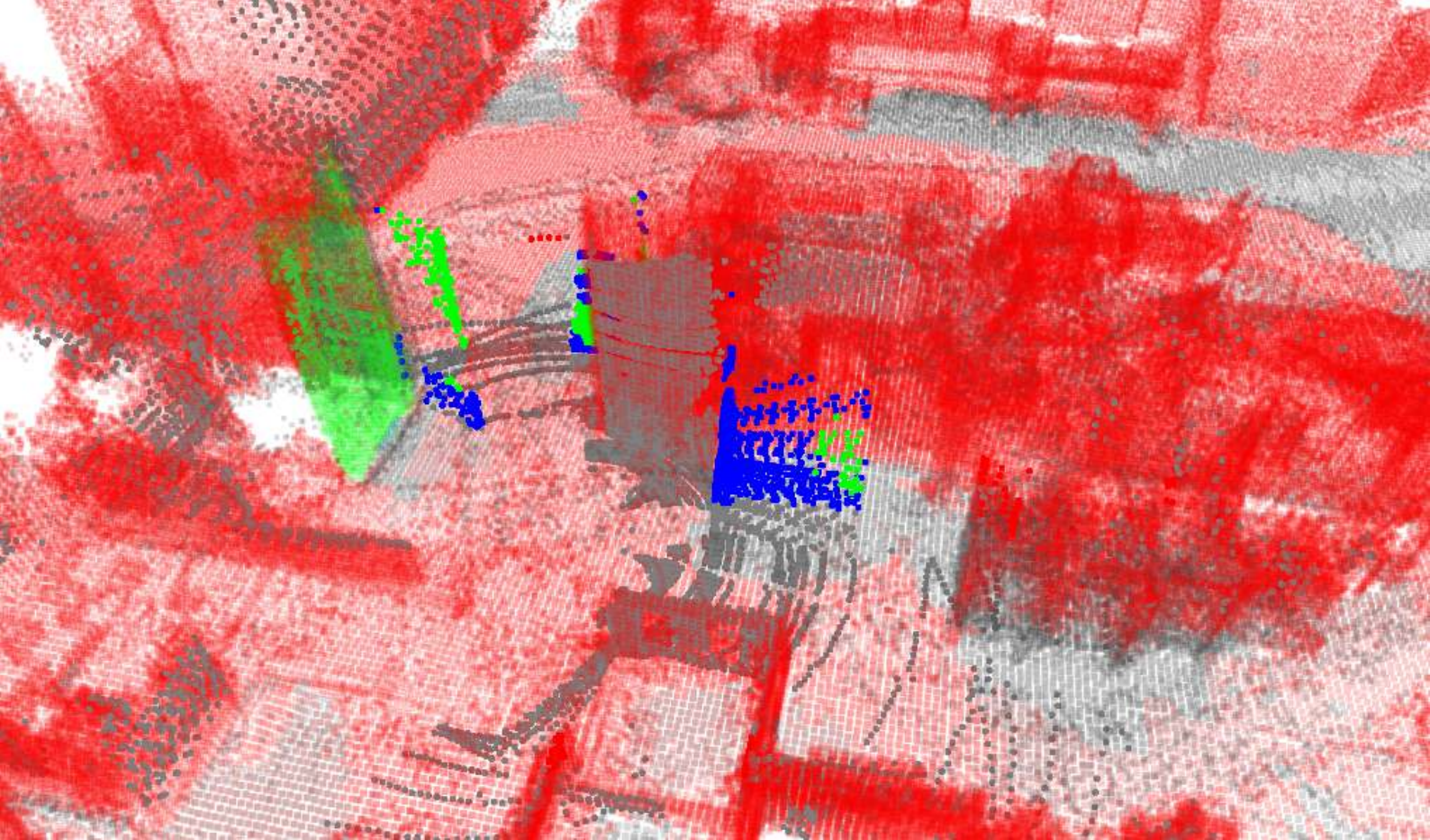}
            \caption{SPS~\cite{hroob2024ral}}
        \end{subfigure} &
        \begin{subfigure}{0.13\textwidth}
            \centering
            \includegraphics[width=\linewidth]{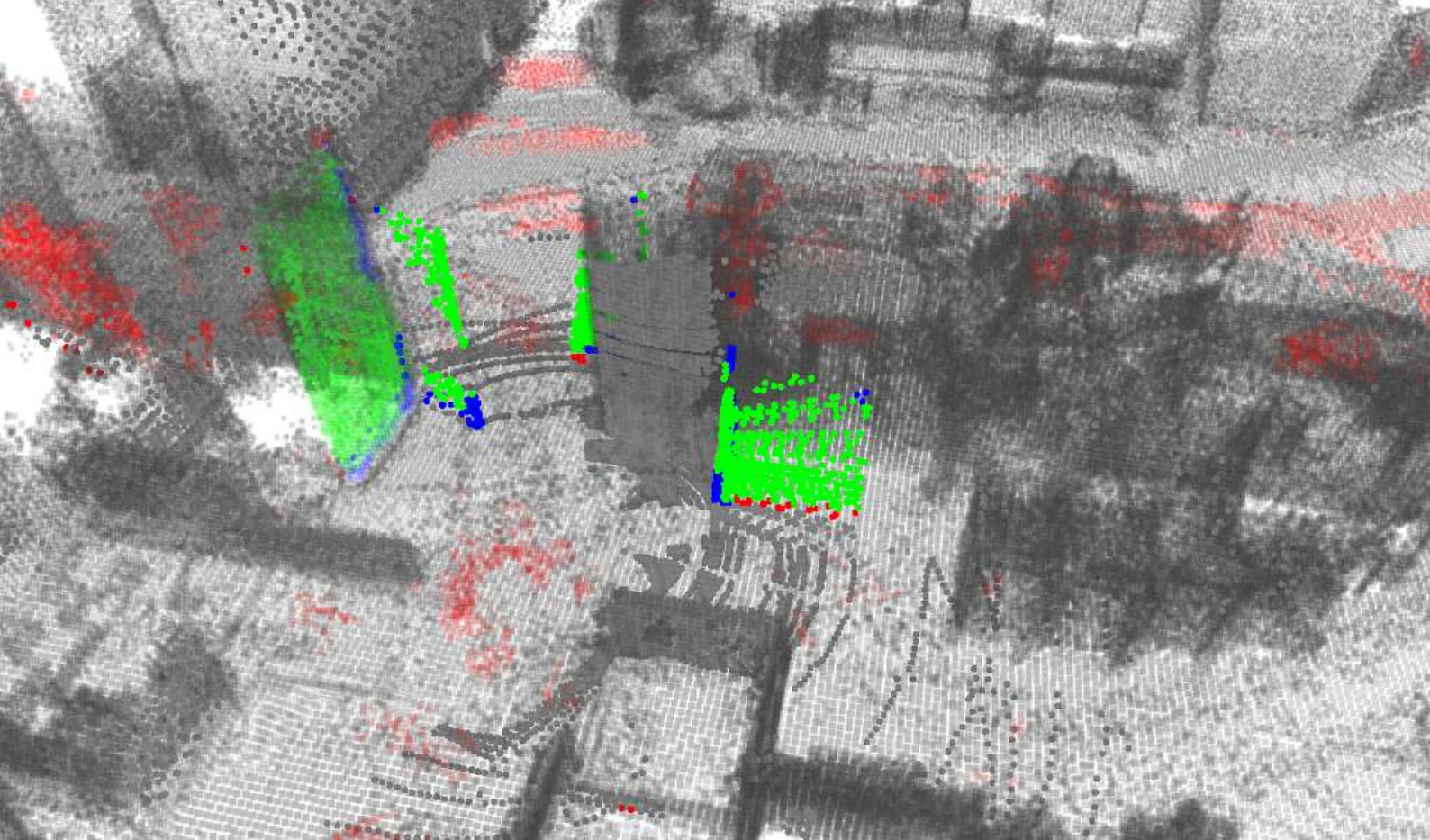}
            \caption{Ours}
        \end{subfigure} \\
    \end{tabular}
    \caption{Qualitative comparison of different change detection methods on our custom dataset. 
    The first column represents the prior map, and the second column corresponds to the current scan data.
    The third to \changed{seventh} columns visualize the results \changed{from other SOTA methods} and our proposed approach, 
    overlaying the predictions onto both the map and the scan data.
    Green, red, and blue points correspond to true \changed{changes}, false \changed{changes}, and false \changed{statics}, respectively.
    The fewer red and blue points, the better the result.}
    \label{fig:custom_dataset_eval}
\end{figure*}

\subsection{Evaluation of Change Detection Performance}\label{subsec:exp_change_detection}
In scan-wise evaluation, visibility and occupancy-based methods recorded the lowest IoU scores across all three sequences, as shown in~\tabref{tab:custom_evaluation}.
As illustrated in~\figref{fig:custom_dataset_eval}(c) and~(d), \changed{both visibility~\cite{underwood2013icra} and occupancy~\cite{walcot2012iros}} methods include a significant number of false change points.
This result demonstrates that geometric threshold-based change detection methods are vulnerable to data occlusion and noise. 
In contrast, deep learning-based approaches exhibit robust performance in scan-wise change detection, as they do not rely on geometric discrepancy.
However, since SPS is purely based on a regression network, its change detection boundaries are vague and highly sensitive to threshold selection. 
It is evidenced by the presence of \changed{false change} points near ground-level regions, as shown in the first row of~\figref{fig:custom_dataset_eval}(f).
In contrast, our method and MapMOS, which use classification-based change detection, outperform SPS in \changed{the} scan-wise IoU.
In the map-wise evaluation, both SPS and MapMOS underperformed compared with geometry-based methods.
SPS computes stability score only in voxels with scan points, ignoring others during training, and leads to inaccurate predictions.
MapMOS also classifies changes only where scan points exist, thus often fails in unobserved regions.
In contrast, our method leverages both map and scan points with cross-visibility, 
and refines the map using a Bayes filter conditioned on visibility confidence, 
thus enabling more accurate change detection by avoiding updates in low-confidence regions.

\begin{figure}[!t]
    \centering
    \captionsetup{font=footnotesize}
        \begin{subfigure}[b]{0.48\textwidth}
            \centering
            \includegraphics[width=1.0\textwidth]{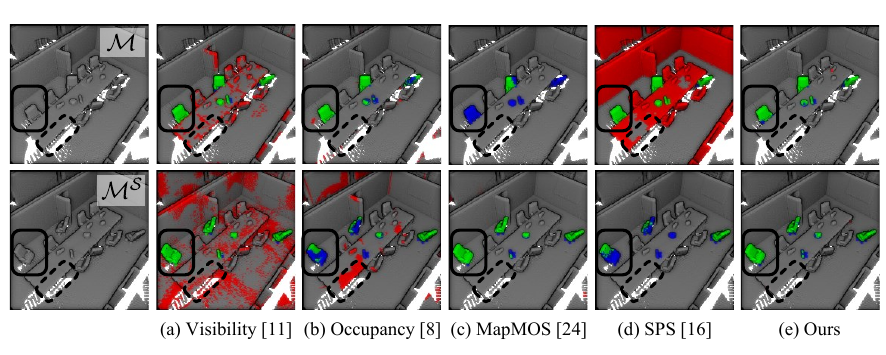}
        \end{subfigure}
    \caption{Qualitative evaluation on the LiSTA dataset. $\mathcal{M}$: prior map, $\mathcal{M^S}$: accumulated current scans.
    The solid line represents areas where the distance between the changed chairs is small, and changes are not correctly detected using distance-based thresholding. 
    The dashed line indicates areas where occlusion frequently occurs in scans due to the sensor's blind spots.}
    \label{fig:lista_qualitative}
    \vspace{-1mm}
\end{figure}

As shown in~\tabref{tab:lista_evaluation} and~\figref{fig:lista_qualitative},~we evaluated the performance of our method {with} the LiSTA dataset.
Similar to the custom dataset, both our method and MapMOS showed robust performance in scan-wise change detection.
In the map-wise evaluation, our method showed higher preservation but a lower rejection rate than geometry-based methods. 
The LiSTA dataset's characteristics, which collects scan data from a limited number of discontinuous positions, 
lead to conservative map updates and lower rejection rates.
However, this limitation is less critical in real-world operations with sufficient observation frequency.

\begin{table}[!t]
    \centering
    \captionsetup{font=footnotesize}
    \caption{Change detection performance comparison on the LiSTA dataset {in terms of scan-wise IoU and map-wise PR, RR, and $\mathrm{F_1}$ scores. Best results in \textbf{bold}, second best in \hl{a gray background}.} }
    \setlength{\tabcolsep}{7pt}
    {\scriptsize
        \label{tab:lista_evaluation}
        \begin{tabular}{llcccc}
        \toprule
        \multirow{2}{*}[-0.5em]{Seq.} & \multirow{2}{*}[-0.5em]{Method}  & \multicolumn{1}{c}{Scan-wise} & \multicolumn{3}{c}{Map-wise} \\ 
        \cmidrule(lr){3-3} \cmidrule(lr){4-6} 
                            &                       & IoU{~$\uparrow$}           & PR{~$\uparrow$}               & RR{~$\uparrow$}              & $\mathrm{F_1}${~$\uparrow$}  \\  \midrule
        \multirow{4}{*}{{\texttt{Simu-1}}}          & Visibility~\cite{underwood2013icra}              & 0.046          & 0.952            & 0.791           & \hl{0.864}           \\
                                                    & Occupancy~\cite{walcot2012iros}                  & 0.484          & 0.723            & {0.934}         & {0.815}  \\
                                                    & MapMOS~\cite{mersch2023ral}                      & \textbf{0.672} & {0.672}          & 0.996           & 0.257           \\
                                                    & SPS~\cite{hroob2024ral}                          & 0.353          & 0.361            & 0.997           & 0.530           \\
                                                    & {Chamelion~(ours)}                               & \hl{0.621}        & {0.901}          & {0.836}         & \textbf{0.867}      \\ \midrule

        \multirow{4}{*}{{\texttt{Simu-2}}}          & Visibility~\cite{underwood2013icra}              & 0.118          & {0.973}          & 0.585           & 0.731           \\
                                                    & Occupancy~\cite{walcot2012iros}                  & 0.618          & 0.743            & {0.946}         & \textbf{0.832}  \\
                                                    & MapMOS~\cite{mersch2023ral}                      & \hl{0.732}        & {0.987}          & 0.085           & 0.156           \\
                                                    & SPS~\cite{hroob2024ral}                          & 0.677          & 0.298            & {0.994}         & 0.458           \\
                                                    & {Chamelion~(ours)}                               & \textbf{0.764}        & 0.916            & 0.761           & \hl{0.831}           \\ \midrule

        \multirow{4}{*}{{\texttt{Simu-3}}}          & Visibility~\cite{underwood2013icra}              & 0.035          & {0.968}          & 0.729           & \hl{0.832}           \\
                                                    & Occupancy~\cite{walcot2012iros}                  & 0.280          & 0.709            & 0.951           & {0.812}  \\
                                                    & MapMOS~\cite{mersch2023ral}                      & {0.307}        & {0.993}          & 0.073           & 0.136           \\
                                                    & SPS~\cite{hroob2024ral}                          & \hl{0.346}     & 0.362            & 0.949           & 0.524           \\
                                                    & {Chamelion~(ours)}                               & \textbf{0.348} & 0.911            & 0.779           & \textbf{0.840}      \\ \bottomrule
        \end{tabular}
        \vsfig
    }
\end{table}

\begin{figure}[!t]
	\captionsetup{font=footnotesize}
	\centering
	\includegraphics[width=0.45\textwidth]{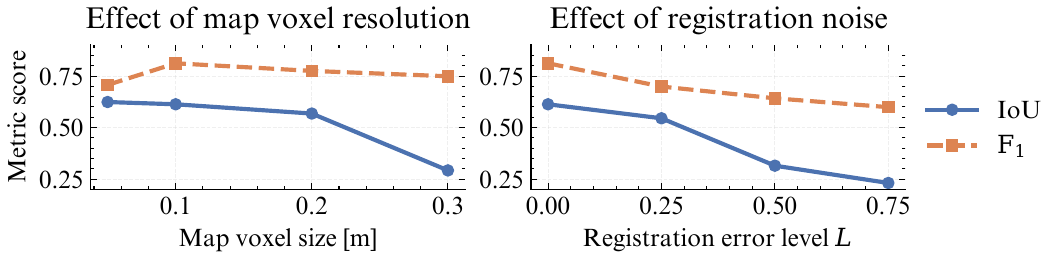}
	\caption{\resubmit{Performance under varying voxel sizes and registration errors.}}
	\label{fig:limitations_pseudo}
	\vspace{-5mm}
\end{figure}
\resubmit{\subsection{Effect of Map Resolution and Registration Errors}\label{subsec:exp_map_update}
Since our pseudo training dataset was based on a single-session indoor environment, 
the map was generated with a small fixed voxel size and under a low-drift condition. 
However, in real-world multi-session environments, the voxel resolution of the map may differ, 
and map-to-scan registration between sessions can be imperfect.
To verify whether the model trained with pseudo-label supervision remains reliable under such conditions,
we conducted two robustness experiments on the custom dataset, as summarized in~\figref{fig:limitations_pseudo}.}

\resubmit{\textbf{Sensitivity to map voxel size.} 
We varied the voxel size of the prior map from 0.05 m to 0.3 m while keeping the scan resolution fixed.
As shown in the results, the map-wise performance peaked at 0.1 m and decreased for other voxel sizes, and the scan-wise IoU was highest at 0.05 m and dropped sharply beyond 0.2 m.
This indicates that the model maintains stable performance when the scan or map density is close to the training voxel size, whereas resolution mismatches reduce geometric consistency.}

\resubmit{
\textbf{Sensitivity to registration errors.}
We further analyzed robustness to inter-session map-to-scan misalignment by injecting synthetic SE(3) noise into the map-to-scan registration matrix~$\mathbf{G}$:
\begin{equation}
\mathbf{G}' = \exp(\hat{\boldsymbol{\omega}}_L)\mathbf{G} + \mathbf{t}_L,
\end{equation}
where $L$, $\mathbf{t}_L \sim \mathcal{N}(\mathbf{0}, L^2\Sigma_T)$, and $\boldsymbol{\omega}_L \sim \mathcal{N}(\mathbf{0}, L^2\Sigma_R)$ denote the perturbation level, 
translation noise, and rotation noise, respectively; and the hat operator $\hatop$ maps the rotation vector to its skew-symmetric form for the exponential map.
The maximum perturbations were bounded by $0.05\mathrm{m}$ in translation, $1.0^{\circ}$ in roll/pitch, and $2.0^{\circ}$ in yaw.
Both IoU and $\mathrm{F}_1$ scores consistently declined as map-to-scan misalignment increased, 
confirming that accurate map-to-scan inter-session registration is essential for reliable pseudo-label supervision.}

\begin{table}[t]
  \centering
  \captionsetup{font=footnotesize}
  \caption{\resubmit{Ablation study on the effect of {high-dynamic} (HD) object removal in scan and map.}}
  {\scriptsize
  \begin{tabular}{lcc|cccc}
      \toprule
      \multirow{2}{*}{ } & \multicolumn{2}{c|}{\textbf{HD removal}} & \multicolumn{4}{c}{\textbf{Performance metrics}} \\
      \cmidrule(lr){2-3}\cmidrule(lr){4-7}
      & {Scan} & {Map} & IoU~$\uparrow$ & PR~$\uparrow$ & RR~$\uparrow$ & $\mathrm{F}_1~\uparrow$ \\
      \midrule
      {[A]} & \ding{55} & \ding{55} & 0.642 & {0.981} & 0.558 & 0.711 \\
      {[B]} & \checkmark & \ding{55} & 0.638 & 0.980 & {0.722} & \textbf{0.831} \\
      {[C]} & \ding{55} & \checkmark & \textbf{0.725} & {0.981} & 0.583 & 0.731 \\
      {[D]} & \checkmark & \checkmark & \textbf{0.725} & 0.980 & {0.722} & \textbf{0.831} \\
      \bottomrule
  \end{tabular}
  \vspace{-1.5em}
  }
  \label{tab:hdremoval_ablation}
\end{table}

\begin{figure}[h!]
    \centering
    \captionsetup{font=footnotesize}
    \includegraphics[width=0.45\textwidth]{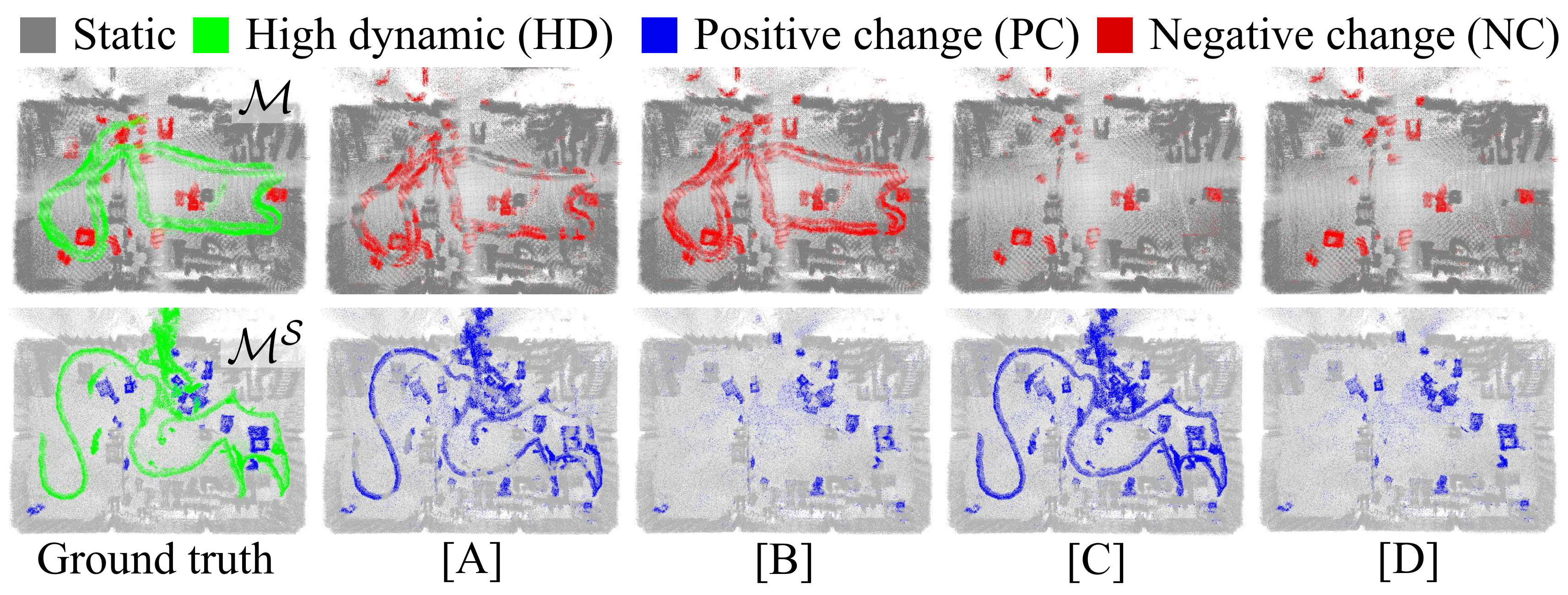}
    \caption{\resubmit{Qualitative visualization of high-dynamic (HD) object removal corresponding to \tabref{tab:hdremoval_ablation}.
    $\mathcal{M}$: prior map, $\mathcal{M^S}$: accumulated current scans.}}
    \label{fig:hd_dataset_information}
\end{figure}
\vsfig
\vspace{-1.5em}
\resubmit{\subsection{Effect of High-Dynamic Objects Removal}\label{subsec:exp_hd_removal}
We evaluate the Const-1F sequence in our custom dataset with and without high-dynamic (HD) object removal to assess its impact on low-dynamic (LD) change detection.
HD points account for 4.6\% of all scan points and 6.46\% of all map points.
As shown in \tabref{tab:hdremoval_ablation}, removing HD objects improves performance:
the map-wise F$_1$ score increases from 71.1\% to 83.1\%, and the scan-wise IoU rises from 63.8\% to 72.5\%.
This suggests that residual HD objects in maps leave temporal artifacts that hinder LD detection.
Although the overall performance remains above 60\%, the results underscore the importance of HD removal for consistent performance.}

\begin{table}[!h]
    \centering
    \captionsetup{font=footnotesize}
    \caption{\resubmit{Runtime comparison on different platforms.}}
    \setlength{\tabcolsep}{10pt}
    {\scriptsize
        \begin{tabular}{lccc}
            \toprule
            Platform  & FPS~(Hz) & IoU~(Scan-wise) & $\mathrm{F}_1$~(Map-wise) \\ \midrule
            RTX 3060  & 14.83 & 0.545 & 0.693 \\
            Orin NX & 2.74 & 0.544 & 0.693 \\
            \bottomrule
        \end{tabular}
    }
    \label{tab:runtime_w_jetson}
    \vspace{-4mm}
\end{table}
\subsection{Runtime}\label{subsec:exp_runtime} 
\resubmit{We benchmark our pipeline on both a desktop GPU (RTX~3060) and an embedded platform (Jetson Orin~NX, 8\,GB). 
As shown in Table~\ref{tab:runtime_w_jetson}, the network runs at 14.8\,Hz on the RTX~3060 and 2.74\,Hz on the Orin~NX. 
Although the embedded device operates at a lower speed, the performance remains consistent. This confirms that GPU-equipped embedded systems can support practical deployment.}

\subsection{Ablation Study}\label{subsec:exp_ablation}
\begin{table}[t]
    \centering
    \captionsetup{font=footnotesize}
    \caption{\resubmit{Evaluation of supervision source across training and test domains. M: manual supervision, P: pseudo supervision.}}
    \label{tab:label_source_ablation}
    {\scriptsize
      \begin{tabular}{l l c c c c}
      \toprule
      {Test} & {Training} & {IoU}~$\uparrow$ & {PR}~$\uparrow$ & {RR}~$\uparrow$ & {F$_1$}~$\uparrow$ \\
      \midrule
      \multirow{2}{*}{Custom} 
        & Custom (M)     & 0.059 & 0.965 & 0.637 & 0.761 \\
        & Custom (P)     & \textbf{0.523} & 0.952 & 0.717 & \textbf{0.814} \\
      \midrule
      \multirow{2}{*}{LiSTA} 
        & Custom (M)     & 0.014 & 0.950 & 0.695 & 0.802 \\
        & Custom (P)     & \textbf{0.512} & 0.821 & {0.778} & \textbf{0.822} \\
      \bottomrule
    \end{tabular}}
    \vspace{1.5pt}
  \end{table}
  
\begin{table}[t]
  \centering
  \captionsetup{font=footnotesize}
  \caption{\resubmit{Evaluation of manual vs. pseudo supervision on the LiSTA dataset. M: manual supervision, P: pseudo supervision.}}
  \label{tab:lista_supervision_ablation}
  {\scriptsize
    \begin{tabular}{l l c c c c}
    \toprule
    {Test} & {Training} & {IoU}~$\uparrow$ & {PR}~$\uparrow$ & {RR}~$\uparrow$ & {F$_1$}~$\uparrow$ \\
    \midrule
    \multirow{2}{*}{LiSTA} 
    & LiSTA (M) & 0.074 & 0.849 & 0.526 & 0.649 \\
    & LiSTA (P) & \textbf{0.334} & 0.603 & 0.705 & \textbf{0.650} \\
    \bottomrule
  \end{tabular}
  }
  \vspace{-3.5pt}
\end{table}

\resubmit{
\textbf{Ablation study on Composition-based Data Augmentation Effectiveness.}
~To evaluate the effectiveness of our data augmentation, we trained the model using two types of training labels: manual and pseudo-labeled data.
As summarized in \tabref{tab:label_source_ablation}, Custom~(M) was trained with 4,700 manually annotated labels, while Custom~(P) used 5,783 pseudo labels generated in \secref{subsec:mono_temporal_datagen}.
Custom~(P) outperformed Custom~(M) in both IoU and $\mathrm{F}_1$ score, 
indicating that composition-based augmentation exposes the model to more diverse environmental variations than manual labeling. 
We further trained the model on LiSTA Simu-1 using two settings: LiSTA~(M) with manual ground truth and LiSTA~(P) with newly generated pseudo labels while deliberately ignoring the available ground truth. 
Both models were then tested exclusively on Simu-2 and Simu-3 sequences.
LiSTA~(P) achieved higher performance than LiSTA~(M) on unseen sequences~(see~\tabref{tab:lista_supervision_ablation}), 
demonstrating that the proposed augmentation generalizes well across datasets.}

\textbf{Ablation study on Network Design.}
To validate our method, we conducted an ablation study in~\tabref{tab:ablation_study} on key aspects:~dual-head structure, and backbone feature division.

We first investigate the impact of using a dual-head structure.
[A] presents a model with only a class head, while [B] and [C] employ a dual-head structure.
[A] yields a significantly lower rejection rate (35.9\%) compared with [B] and [C] (49.8\% and 71.5\%, respectively).
Without confidence estimation, the network continuously updates incorrect predictions in occluded regions, 
causing false statics to accumulate and biasing the output toward the static class.

Next, we compare confidence heads trained with \changed{only} high-level features~[B] and \changed{only} low-level features~[C].
Although semantically rich, high-level features lack geometric \changed{details}, leading to overconfident predictions and incorrect updates in uncertain regions.
This results in widespread misclassification toward the static class.

\begin{table}[!t]
    \centering
    \captionsetup{font=footnotesize}
    \caption{Ablation study on our composition-based data augmentation, dual-head architecture, and backbone feature division. Best results in bold.} 
    {\scriptsize
    \label{tab:ablation_study}
    \begin{tabular}{lcc|cccc}
    \toprule
    & \multirow{2}{*}[-0.5em]{\textbf{Dual head}} & \multirow{2}{*}[-0.5em]{\textbf{Feat.~div.}} & \multicolumn{4}{c}{\textbf{Performance metrics}} \\ 
    \cmidrule(lr){4-7}
    &  & \multirow{2}{*}{} &  IoU{~$\uparrow$}  & PR{~$\uparrow$}  & RR{~$\uparrow$}  & $\mathrm{F_1}${~$\uparrow$} \\
    \midrule
    {[A]} & \ding{55}    & \ding{55}             & 0.373           & 0.995           & 0.359           & 0.527 \\ 
    {[B]} & \checkmark   & \ding{55}             & 0.385           & 0.948           & 0.498           & 0.645 \\ 
    {[C]} & \checkmark   & \checkmark            & \textbf{0.613}           & 0.949           & 0.715           & \textbf{0.812} \\ 
    \bottomrule
    \end{tabular}
    \vsfig
    }
\end{table}

\begin{figure}[t!]
    \centering
	\captionsetup{font=footnotesize}
    \begin{subfigure}[b]{0.46\textwidth}
        \includegraphics[width=1.0\textwidth]{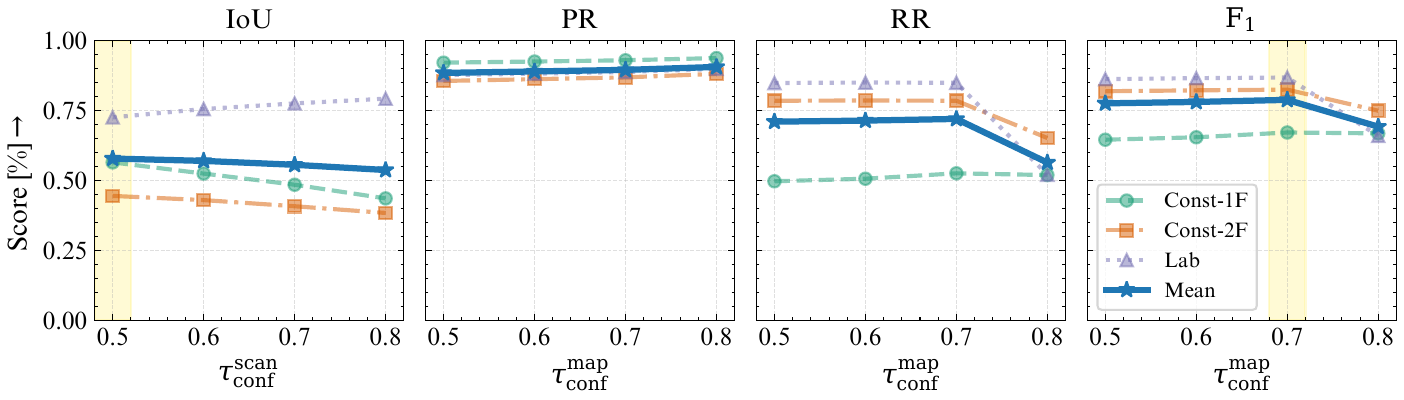}
        \caption{}
    \end{subfigure}
    \begin{subfigure}[b]{0.46\textwidth}
        \includegraphics[width=1.0\textwidth]{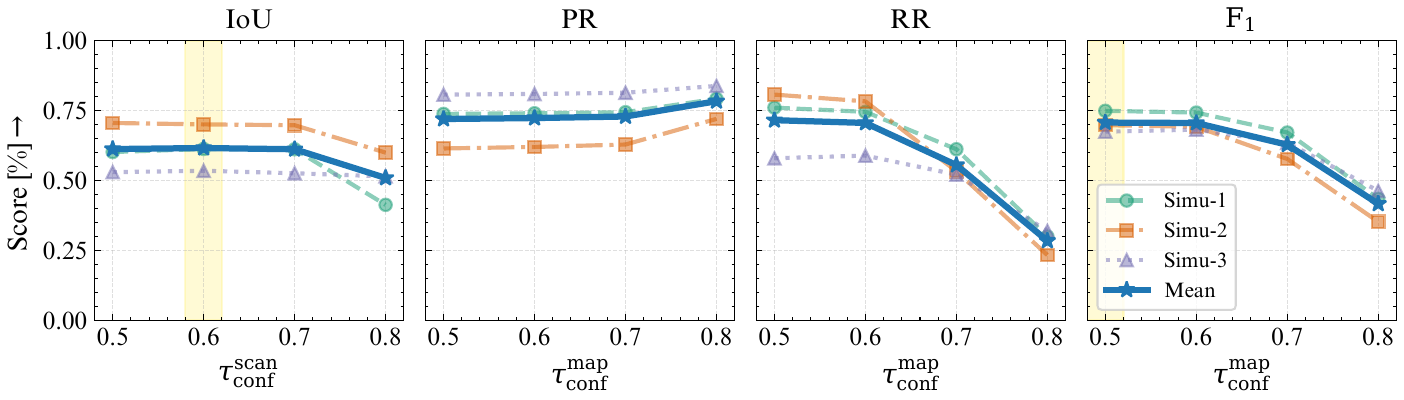}
        \caption{}
    \end{subfigure}
    \caption{
    Ablation study of the confidence threshold~$\tau_{\mathrm{conf}}$ on the (a)~Custom and (b)~LiSTA datasets. 
    We report scan-wise IoU and map-wise PR, RR, and~$\mathrm{F}_1$ score under different threshold values. 
    Yellow shaded regions indicate the recommended range of~$\tau_{\mathrm{conf}}$ for stable performance.
    }
    \label{figure:ablation_tau}
    \vspace{-5mm}
\end{figure}

\resubmit{\textbf{Ablation study on Confidence Threshold.}~Since our pipeline updates changes with visibility confidence, 
we evaluate the sensitivity of performance to the confidence threshold~$\tau_{\mathrm{conf}}^{\mathrm{scan}}$ and $\tau_{\mathrm{conf}}^{\mathrm{map}}$. 
Specifically, we sweep each threshold over $\{0.5, 0.6, 0.7, 0.8\}$ on the custom and LiSTA datasets, 
and report scan-wise intersection-over-union~(IoU) and map-wise preservation rate~(PR), rejection rate~(RR), and~$\mathrm{F}_1$ score.
See~\figref{figure:ablation_tau} for the detailed results.}
\resubmit{
  For the real-world custom dataset, noise and registration drift made a lower threshold ($\tau_{\mathrm{conf}}^{\mathrm{scan}}=0.5$) effective for suppressing false positives in scans, 
  while higher threshold ($\tau_{\mathrm{conf}}^{\mathrm{map}}=0.7$) were better for maps.
  In contrast, in the simulated LiSTA dataset, with no drift and generally higher confidence, achieved the best scan IoU at $\tau_{\mathrm{conf}}^{\mathrm{scan}}=0.6$.
  Because it contains fewer scans and less frequent updates than the custom dataset, setting $\tau_{\mathrm{conf}}^{\mathrm{map}}=0.5$ for maps expanded the update regions and improved performance.
}

\vspace{-0.3cm}
\section{Conclusion}
In this paper, we introduced {\acro}, a novel approach for online change detection and long-term 3D map management.
Our dual-head network structure is designed to robustly detect changes, even with occlusion, 
and it can be trained using pseudo change datasets generated by composition-based augmentation.
Moreover, our method exhibits robust performance across various environmental settings, 
demonstrating its generalization \changed{capability}. 
\resubmit{Currently, our framework learns a fixed cross-visibility from pseudo-labeled datasets 
and fuses change classification results based on it. Consequently, it performs better when registration errors are small or map voxel sizes are similar to the training data. 
As future work, we plan to dynamically compute cross-visibility to enhance robustness under varying registration errors and voxel resolutions.}


\bibliographystyle{URL-IEEEtrans}
\vspace{-0.3cm}
\bibliography{URL-bib}


\end{document}